\documentclass[twoside,11pt]{article}

\usepackage[nonatbib,preprint]{jmlr2e}

\usepackage[utf8]{inputenc} % allow utf-8 input
\usepackage[T1]{fontenc}    % use 8-bit T1 fonts
\usepackage{hyperref}       % hyperlinks
\usepackage{url}            % simple URL typesetting
\usepackage{booktabs}       % professional-quality tables
\usepackage{amsfonts}       % blackboard math symbols
\usepackage{nicefrac}       % compact symbols for 1/2, etc.
\usepackage{microtype}      % microtypography
\usepackage{wrapfig}
\usepackage{bbm}
\usepackage{tikz}
\usetikzlibrary{shapes.geometric, arrows}
\tikzset{
	treenode/.style = {shape=rectangle, rounded corners,
		draw, align=center,
		top color=white, bottom color=blue!20},
	root/.style     = {treenode, font=\Large, bottom color=red!30},
	env/.style      = {treenode, font=\ttfamily\normalsize},
	dummy/.style    = {circle,draw}
}

\usepackage[cmex10]{amsmath}
\usepackage{amsthm}
\DeclareMathOperator*{\argmax}{argmax}

\usepackage{bm}
\usepackage{paralist,xparse}
\usepackage{dsfont}
\usepackage{amsfonts}
\usepackage{multirow} %Multi-Row Stuff
\usepackage{subfigure}
\usepackage{amssymb} %Math Symbols
\usepackage{scrextend}
\usepackage{float}
\usepackage{adjustbox}

\usepackage{algorithm}
\usepackage{algorithmic}
\usepackage{mathtools}
\usepackage{times}
\usepackage{pgfplots}
\usepackage{array}
\usepackage{tikz}
\usepackage[european resistor, european voltage, european current]{circuitikz}
\usetikzlibrary{arrows,shapes,positioning,snakes}
\usetikzlibrary{decorations.markings,decorations.pathmorphing,decorations.pathreplacing}
\usetikzlibrary{calc,patterns,shapes.geometric,topaths}

\newtheorem{theorem}{Theorem}
\newtheorem{definition}{Definition}

\newtheorem{corollary}{Corollary}

\newcommand{\norm}[1]{\lVert#1\rVert} %For norm Command
\newcommand{\E}{\Bbb{E}}

\ShortHeadings{Bayesian Revealed Preferences for YouTube Commenting Behavior}{Hoiles, Krishnamurthy and Pattanayak}
\firstpageno{1}

\begin{document}

\title{Rationally Inattentive Inverse Reinforcement Learning Explains YouTube Commenting Behavior}

\author{\name William Hoiles \email  william.hoiles@katerra.com \\
       \addr Director of AI and Cloud Infrastructure\\ KATERRA\\Menlo Park, CA $94025$, USA
       \AND
       \name Vikram Krishnamurthy \email vikramk@cornell.edu \\
       \addr Electrical and Computer Engineering\\
       Cornell University\\
       Ithaca, NY $14853$, USA
       \AND
       \name Kunal Pattanayak \email kp487@cornell.edu \\
       \addr Electrical and Computer Engineering\\
       Cornell University\\
       Ithaca, NY $14853$, USA
       }

%\editor{Kevin Murphy and Bernhard Sch{\"o}lkopf}

\maketitle

\begin{abstract}  We consider a novel application of inverse reinforcement learning with behavioral economics constraints to model, learn and predict the commenting behavior of YouTube viewers. Each group of users is modeled as a rationally inattentive Bayesian agent which solves a contextual bandit problem. Our methodology integrates three key components. First, to identify distinct commenting patterns, we use deep embedded clustering to estimate framing information (essential extrinsic  features) that clusters users into distinct groups.
Second, we present an inverse reinforcement learning algorithm that uses Bayesian revealed preferences to test for rationality: does there exist a utility function that rationalizes the given data, and if yes, can it be used to predict commenting behavior?
	%is the agent a utility maximizer, and does an information cost function exist that  rationalizes the data? 
Finally, we impose behavioral economics constraints stemming from rational inattention to characterize the attention span of groups of users.
The test  imposes a R{\'e}nyi mutual information cost constraint which  impacts how the agent can select attention strategies to maximize their expected utility.
After a careful analysis of a massive YouTube dataset, our surprising result is that in most YouTube user groups, the commenting behavior is consistent with optimizing a Bayesian utility with rationally inattentive constraints. The paper
also highlights how the rational inattention model can accurately predict commenting behavior.
The massive YouTube dataset and analysis used in this paper are available on GitHub and completely reproducible.
%The test provides constructive set valued estimates of the utility function and information acquisition  cost  of the agent. We illustrate  these methods on  a massive YouTube dataset for characterizing the commenting behavior of users - remarkably, the commenting behavior of majority of the groups fits the rationally inattentive utility maximization model.
\end{abstract}


\section{Introduction}
\label{sec:Introduction}
%\textcolor{red}{add that this is convex feasibility}
%\textbf{\textcolor{blue}{This introduction reads like scrambled eggs - it repeats things in multiple paragraphs; we need a more careful organization: 
		%Have one paragraph to say that this paper deals with a novel application of  inverse RL in YouTube. 
		%THen have a second paragraph discussing revealed preferences; then a third paragraph describing the rational inattention behavioral constraint; then a forth paragraph explaining deep clustering, finally a paragraph explaining the relevance of YouTube inverse RL, commenting behavior, etc. }}
%\textcolor{red}{put category dissection explanation, numbers  of comments (users), sentiment prediction in addition to viewcount, predicting groups of users behavior, clusters - term for partitions, change in terminology over the paper, sort out abstract, section on prediction, section on kkt conditions, utility of ?? needs to be addressed}
%\textcolor{red}{To add: robustness results for the cases that go through}
This paper considers a novel application of inverse reinforcement learning with behavioral economics constraints to model, learn and predict the commenting behavior of YouTube viewers. We model each group of users as a Bayesian rationally inattentive agent. (Equivalently, each agent solves a contextual bandit problem with behavioral economics constraints.) Our methodology integrates three key components to model the collective commenting behavior of YouTube viewers. First, to identify distinct commenting patterns over YouTube videos, we use \textit{Deep Embedded Clustering} that partitions groups of videos from a massive YouTube dataset into non-overlapping segments.
% called "frames". 
For each segment, we define a group comprising the users that view and comment on videos within that segment- each group's behavior is individually analyzed. The second component involves inverse reinforcement learning; we use \textit{Bayesian Revealed Preferences} to construct a  test for utility maximization behavior for individual groups of users and estimate utility functions that rationalize their behavior. Finally, we impose behavioral economics constraints stemming from \textit{Rational Inattention} to characterize the attention span used by the viewers while commenting. This rationally inattentive model of commenting behavior can be interpreted as a  constrained contextual bandit problem as will be discussed in Sec.\ \ref{sec:ProblemFormulationandRationalInattention}. After a careful analysis of a massive
YouTube dataset, our surprising result is that in most YouTube user groups, the overall commenting behavior\footnote{By overall commenting behavior in YouTube, we mean both comment count and video ratings (likes and dislikes). Another term used in the literature (\cite{KH17}) is ``user engagement''. This paper characterizes the overall user engagement in a YouTube dataset.}
is consistent with optimizing Bayesian utility with rationally inattentive constraints. The paper
also highlights how the rational inattention model can predict commenting
behavior.
%After a careful analysis of a massive YouTube dataset, our surprising result is that in most YouTube user groups, the commenting behavior is consistent with optimizing a Bayesian utility with rationally inattentive constraints. The paper also highlights how the rational inattention model may also help in predicting future commenting behavior.
%%%%%%%%%%%%%%%%%%%%%%%%%%%%%%

%This paper considers a novel application of Bayesian inverse reinforcement learning with behavioral economics constraints to explain/predict the commenting behavior of YouTube users. We show the surprising result that in many cases, the commenting behavior of YouTube users is consistent with optimizing a Bayesian utility with rationally inattentive constraints.
To better explain the main ideas, consider the following abstract setup: suppose  a Bayesian agent chooses an action to maximize an expected utility function based on the noisy measurement of an underlying state. Assume that the Bayesian agent is rationally inattentive, that is, obtaining this noisy measurement is expensive  -- 
this information acquisition cost affects the action chosen by the agent.  An observer (analyst) records the dataset of actions of the Bayesian agent and knows the underlying state. 
%The aim of this paper is to solve the following inverse reinforcement learning problem: How can the observer estimate the utility function that is consistent with the given dataset dataset, constrained to usage of a particular information acquisition cost structure? 
%Equivalently, the objective is to obtain set valued estimates of utility and information acquisition cost that are consistent with the  dataset.
Classical inverse reinforcement learning aims to estimate the utility function of a decision process by observing its decisions and stimulus input~(\cite{NG00})  while assuming the agent is a utility maximizer.  \cite{FU17} discuss construction of disentangled state dependent rewards for agents.
  %Our work produces analogous results, albeit through a different methodology whose computation time depends on feasibility testing of a mixed integer linear program.
In such problems, the existence of a utility function the agent maximizes is assumed implicitly. The revealed preference approach in this paper addresses a deeper and more fundamental question: does a utility function exist that rationalizes the given data (with rational inattention constraints) and if yes, how to estimate this utility function.
%Furthermore, does the agent's attention span (information acquisition cost) conform to a particular structure. 
%The methodology that we use outlines ways to construct a set valued utility function if it exists, while commenting on the type of rationality an agent possesses. 

%We also use these estimates on a massive YouTube data set to model  (and predict) the commenting
%behavior of users.   %ideos become popular over time.
%The rational inattention theory with costly information acquisition presented in this paper facilitates learning the utility function of an agent by  observing their action selection policy.
%This  can be viewed as a methodology for solving the temporal credit assignment problem  in preference-based inverse reinforcement learning~\cite{WANF17}.
%\footnote{Popularity  prediction of multimedia is important in designing adaptive caches in 5G wireless communications, social media modeling and more generally understanding human interaction with social media.} over time. 

%Our methodology  stems from behavioral economics and deep embedded clustering:  non-parametric estimation of utility functions and feature extraction using deep clustering to construct behavioral economics based models  for  Bayesian agents. Let us briefly explain these two aspects.
Estimating  utility functions given a finite length time series of decisions and budget constraints is widely studied in the area of revealed preferences  in economics, starting with the paper of \cite{AF67} which gives a remarkable necessary and sufficient condition for utility maximization; see also \cite{VR82,Var12,Woo12} and more recently  in machine learning~(\cite{LO09}).
\cite{VR83} characterized investor behavior by devising Afriat-type conditions for expected utility maximization in a Bayesian framework.
 Unlike~\cite{AF67} and \cite{VR83}, the utility function in our Bayesian set-up involves discrete valued variables. 
 
 Costly  information acquisition by Bayesian agents has been studied by  economists and psychologists under the area of ``rational inattention'' pioneered by Nobel Laureate Christopher Sims~(\cite{Sim03,Sim10}). Rational Inattention is a form of bounded rationality - the key idea is that human attention spans for information acquisition are limited and  can be modeled in information theoretic terms as a Shannon capacity limited communication channel. Sim's rational inattention model is studied extensively in behavioral economics~(\cite{MM15}). \cite{Woo12} considered an upper bound on the Shannon capacity  for testing  rational inattention with visual perception queues. Typically, the information acquisition  costs faced by a decision maker are not known to the observer (analyst). A general test for rational inattention is proposed in two remarkable papers by~\cite{CM15} and \cite{CD15} with minimal restrictions on the  information acquisition cost- we will refer to this as the general cost. 

\subsection*{Framing and Categories}Our analysis involves partitioning the massive YouTube dataset into segments using two different methodologies. The first method constructs non-overlapping segments of videos based on the nature of their framing information gathered from the videos' thumbnail and description. In the second method, the videos are grouped based on the video category (one of the attributes in YouTube data)- all videos falling under the same category constitute a single segment. We shall use the term "frame" and "category" to refer to segments when analyzing data partitioned in the first and second way respectively. 

Let us briefly explain these two methods:
In behavioral economics, \cite{TK81} use ``frames''  to describe information  an agent has when making a decision. 
Regarding analysis of YouTube and other social media datasets, extensive  studies  (\cite{Kha17,HG17,ABCH15,HAK17}) show  that comments  posted by users are  influenced by the thumbnail, title, category, and perceived popularity of each video. \cite{LG07} shows that sharing and circulation of videos is a manifestation of social relationships among youth, thus linking human cognitive behavior to user engagement dynamics. For example, when selecting which product to purchase on a website, the positioning of the products and surrounding content on the website  impacts how humans select a product. Given such external information (image/text/numeric) in which the decision problem is embedded, how can one construct a tractable feature set? We develop deep embedded clustering methods to construct the  frames to  test for rational inattentive agents. The deep embedded clustering  is based on~\cite{XGF16,GGLY17}, however we  design the input, encoder, and decoder to account for the visual perception of the frame of the decision problem which includes image, text, and numeric information. 

%\textcolor{red}{add justfication for category dissection method.}
Commenting behavior in YouTube has been well-studied in literature. \cite{SD10} 
analyze comments and their ratings in various categories of YouTube videos. They illustrate the increasing variance of comment ratings and sentiments with polarizing content in different categories of videos. Using open source sentiment analysis tools, the authors also predict the anticipated comment ratings (community feedback) on unrated comments in YouTube videos. %test if commenting ratings and sentiment across different video categories are statistically different. 
%Their findings indicate that different video categories attract different types of users, thereby generating different commenting behavior, with the "Music" category having the highest ratings and positive comments. 
\cite{SC13} define comment classes to analyze YouTube comments on videos over various categories. They conclude that the distribution of comments over comment classes differs with video category. %Furthermore, \cite{SC13} also analyze the relationship between likes/dislikes and the distribution of comments over various comment classes in videos of each video category.
% Following a similar approach, we analyze YouTube data partitioned with respect to their video category and comment whether user groups corresponding to each video category are rationally inattentive bayesian or not.
\cite{SV14} predict YouTube comment types by classifying them using existing comments in contrast to our case, where we predict the commenting behavior using the videos' content.

%We thus analyze YouTube data considering 
To summarize, the two methodologies we consider, namely frame and category, constitute two distinct ways of partitioning groups of videos in YouTube data - a \textit{user-driven approach} based on how the user infers the videos and a \textit{content-driven approach} based on the video category. The perceived popularity of the underlying state is indicated by each video's viewcount, and the nature of commenting behavior is characterized by the number and sentiment (captured by likes and dislikes for each video) of the comments. 
The two significant extensions considered in this paper are the effects of framing (determined using deep embedded clustering) and video category and the use of R{\'e}nyi mutual information cost for testing rational inattention.
Our rational inattention test  is equivalent to solving the temporal credit assignment problem  in \textit{preference-based inverse reinforcement learning}~(\cite{WANF17}). Such inverse reinforcement learning
is used with non-numeric feedback~(\cite{WFN16}), e.g.\ in socially adaptive path planning~(\cite{KP16,HMARD17}) for robots.
As will be discussed in Sec.\ \ref{sec:ProblemFormulationandRationalInattention},  the methodology in this paper is equivalent to inverse  reinforcement learning of a contextual bandit with rational inattention (behavioral economics) constraints.

%Such a framework has potential to present innovative insights when employed in social media applications. In this work, we highlight how our methodology efficiently captures user behavior in one such popular social multimedia platform, YouTube.
%Regarding analysis of YouTube and other social media data sets, extensive  studies  (\cite{Kha17,HG17,ABCH15,HAK17}) show  that comments  posted by users are  influenced by the thumbnail, title, category, and perceived popularity of each video. \cite{LG07} shows that sharing and circulation of videos is a manifestation of social relationships among youth, thus linking human cognitive behavior to user engagement dynamics.
%\textcolor{blue}{Sketch: Characterization of attention span in other ways -bottom up control, top down control of attention - saliency detection- sequential model - can be shown that our model encapsulates both static and sequential behavior- anticipatory decision making - not suitable for YouTube case-link between Ng and NIAS  }
\subsection*{Main Conclusions on YouTube Dataset} This paper focuses on an application of inverse reinforcement learning with behavioral economics constraints to predict overall YouTube commenting behavior. So to better motivate the paper, we summarize our main findings (see Sec.\ 6 for details). 
%\textcolor{blue}{Our analysis of a massive YouTube dataset shows that the commenting behavior given by the number and sentiment of the comments on YouTube videos is a result of the framing information, state, and decision problem faced by the YouTube user groups modeled as Bayesian agents.} 
Based on  extensive data analysis on
%(\textcolor{red}{pending}) 
groups of YouTube users, where each group consists of approximately $3500$ viewers,  our main take-home message (from a behavioral economics point of view)  is  that YouTube user groups are rationally inattentive in their commenting behavior  and that users prefer to comment on videos that are perceived to be popular. In YouTube, video viewcount is the  independent quantity which governs the commenting behavior since videos need to be viewed first before users can comment or rate the video. Our analysis thus also sheds light on the way videos are "perceived" by groups in YouTube i.e.\ how does the collective sentiment behind popular and non-popular videos vary over different video segments. The set valued utility estimates of commenting behavior for user groups also help in predicting future behavior of the users in each video segment; see Sec.\ \ref{sec:RationalInattentionandUtilityMaximizationintheYouTubeSocialNetwork} for additional conclusions. Our pre-processing uses state-of-the-art NLP tools like GloVe~(\cite{PSM14}) to extract information from videos' text descriptions and combines it with {\em Deep Embedded Clustering} to process videos' metadata for partitioning the YouTube dataset.
The results presented in Sec.\ \ref{sec:RationalInattentionandUtilityMaximizationintheYouTubeSocialNetwork} are completely reproducible, with the  code and datasets publicly available on a GitHub repository.

 %%%%%%add later , that cases that go through need a lot of pertubration to be disturbed%%%%%%%%%%%%
%The rationally inattentive model is also robust for all discussed parameterized information cost structures;

\subsection*{Why Analyze the Commenting Behavior in  YouTube?}
%The main objective of our analysis is to effectively characterize user commenting behavior in YouTube. 
YouTube is clearly a social media site, however, YouTube is also a social
networking site. Classical online social networks are dominated by user-user interactions.
However YouTube is unique in that the interaction between users includes video content–that is,
the interaction follows users-content-users.
Examples where our analysis of YouTube commenting behavior are relevant include:
\begin{compactenum}
\item {\em Behavioral Psychology-} Understanding and modeling how humans perceive social multimedia information in the context of framing and rational inattention~(\cite{PG10, TT11}).
\item {\em Economics of YouTube-} Knowledge of commenting behavior characteristics allows YouTube partners (companies like BroadbandTV Corp. (BBTV) and VEVO) to adapt their user engagement strategies to generate higher viewcounts and increase revenue~(\cite{HAK17}).
\item \textit{Content Caching in  $5$G-} By caching highly popular content at base stations (BS), user demand
for the social media video content can be served locally. This reduces
the overall network traffic and improves the QoE for users
requesting content~(see~\cite{HNK15} and references therein).
\item \textit{YouTube Recommendation-} The variance of the utility function with respect to the optimal action selection policy is a useful measure of uncertainty involved with a Bayesian decision maker~(\cite{LV79,MNR11}). For the YouTube dataset, we found that this variance is higher for categories like Sports, News and lower for less controversial categories like Automobiles, Education. These are in line with results in~\cite{SD10} which reports a higher variance of comment ratings in polarizing topics compared to neutral topics. This variance appears in our finite sample analysis in Appendix~\ref{appdx:Risk-AwareUtilityMaximizationbyOff-PolicyEstimation} which also discusses a method to tune online recommendations for rationally inattentive user groups to maximize their expected utility.
\end{compactenum}

\subsection*{Context and other Applications} 
Since Bayesian revealed preferences with rational inattention constitute the main methodology of this paper, below we give some additional insight.

The rationally inattentive model of the decision-making Bayesian agent belongs to a larger class of decision theories based on stochastic information sampling. Models in this class include: {\em $(1)$~Rational models}~(\cite{SMN55,MCH78}) where information acquisition costs are ignored and agents have time invariant preferences over their actions. {\em $(2)$~Bounded rationality models}~(\cite{CD15,Sim03}) where the Bayesian agent incurs a cost for accurate perception and generates noisy samples conditional on the state (either one-shot or sequentially over time) to take an action for maximizing net expected utility. {\em $(3)$~Evidence accumulation models}~(\cite{KJ10,RC04}) where consumers' attention is modeled by drift-diffusion models that accumulate evidence based on whether they are fixating their gaze on either the product or its price. The decision is taken when any of the alternatives' evidence threshold level is achieved. {\em $(4)$~Parallel constraint satisfaction models}~(\cite{GCK08,MC89}) which assume that information is screened sequentially to highlight salient alternatives and final choice is made when the decision maker reaches sufficient internal coherence.

In contrast, examples of visual attention models relevant to our YouTube context from existing literature in psychology include: {\em ($1$) Top down control of attention}~(\cite{YB13,HH03}) where agents focus on visual aspects relevant to the decision objective  {\em ($2$) Bottom up control of attention}~(\cite{FT10,JD12}) where the agent implicitly constructs a topographic saliency map of the visual to guide attention selection.

Testing for Bayesian utility maximization and recovering the resulting utility function also has applications in finance and online marketing. 	
\begin{compactenum}
\item {\em Online Marketing: }\cite{LE09} show that using $9-$ending price displays of online commodities affects consumer choices which were also found to be consistent with rational inattention.  \cite{HM15} test whether online consumer choices are consistent with limited attention models. They exploit their results to enhance sales of a particular brand by intelligent product recommendation to the online customers. Using a Bayesian framework and Shannon entropy, \cite{MN17} shows how rational inattention of online buyers can be exploited for strategic pricing of goods by online sellers.
\item {\em Finance: }\cite{SR00} provides evidence that investors are boundedly rational with limited information-processing capability. \cite{HN07} show that costly information acquisition of economic news by Bayesian investors results in their portfolio management being myopic with respect to parameters like news frequency and accuracy. They also characterize desired news frequencies for investors depending on whether they are risk-averse or risk-seeking. This characterization can be used by online financial services companies to selectively advertise economic news to their clients on their online platforms.   
\end{compactenum}

\subsection*{Organization}  Sec.\ \ref{sec:ProblemFormulationandRationalInattention} introduces the problem formulation and connection to constrained contextual bandits.
Sec.\ \ref{sec:FrameingInformationDeepLearning} discusses a deep embedded clustering algorithm for associating the observed agent's action  to specific frames. %This ensures that framing effects are taken into account when testing for rational inattention, and constructing the  utility function(s) of the agent.
%, and for policy optimization in each associated frame.
In Sec.\ \ref{sec:DecisionTestforRationalInattentionandRecoverabilityofAgentPreferences} and~\ref{sec:MutualInformationMeasuresandAttentionCost}, the decision test for rational inattention with behavioral economics constraints stemming from R{\'e}nyi mutual information are provided. The decision tests are constructive: they provide estimates
of the  utility function, information acquisition  cost, and attention function.
% , and choice function that rationalize the dataset.
Sec.~\ref{sec:RationalInattentionandUtilityMaximizationintheYouTubeSocialNetwork} applies the methods  to a massive  YouTube dataset to  characterize the commenting behavior of users. Appendix~\ref{appdx:proofs} contains the proofs for our main theoretical results Theorem~\ref{thrm:decisiontestrationalinattention} and Theorem~\ref{thrm:reynicost}. Appendix~\ref{appdx:Risk-AwareUtilityMaximizationbyOff-PolicyEstimation} provides  Bernstein based  finite sample performance bounds.
Appendix~\ref{appdx:DeepClustering} summarizes the implementation details of the deep classifier. Appendix~\ref{apdx:youtube} provides additional details about the number of users comprising YouTube user groups. Finally, Appendix~\ref{apdx:AttentionChoice} deals with estimating the agent's attention and choice functions.

\section{Problem Formulation and Rational Inattention}
\label{sec:ProblemFormulationandRationalInattention}
We first describe the problem formulation from the  point of view of the  rationally inattentive Bayesian agents; and then from the point of view of the observer (data analyst) that views the dataset generated by the agents. Despite our abstract formulation, the reader should keep in mind  the YouTube context outlined above, namely that the rationally inattentive Bayesian agents are groups of YouTube users (this is made precise in Sec.\ \ref{sec:RationalInattentionandUtilityMaximizationintheYouTubeSocialNetwork}), while the observer (data analyst) analyzes the data to test if YouTube user groups are rationally inattentive in their commenting behavior. 
%Also, determine the utility function which quantifies the agents' state dependent preference ordering of actions which rationalizes their behavior. Results from our decision tests are illustrated in great detail in Sec.\ \ref{sec:RationalInattentionandUtilityMaximizationintheYouTubeSocialNetwork}. Note that the decision test (Theorem~\ref{thrm:decisiontestrationalinattention}) to recover utility functions stays the same, while the output may vary with segments of videos; this is expected since distinct user groups perceive and interact with the videos differently}.
%\vspace{-0.25cm}
\subsubsection*{Viewpoint 1. Rationally Inattentive Bayesian Agent}
Assume the agent knows the finite state space  $\mathcal{X}$ and finite action space $\mathcal{A}$. The agent's prior beliefs of the possible states are given by the prior probability distribution $\mu(x)$, $ x\in \mathcal{X}$.
% The agent has access to a measurement device that allows a refined estimate of the state for a given cost that is known to the agent.
The \textit{attention function} $\alpha(s|x)$ of  the agent  provides a distribution over the signals $s\in\mathcal{S}(\alpha)$ when the state is $x$. The set of possible signals $\mathcal{S}(\alpha)$ for a given attention function $\alpha$ is finite. The attention function encodes all the information (signals, private information, and measurement mechanism) available to the agent to compute the   posterior state distribution. Given the prior  $\mu(x)$, and attention function $\alpha(s|x)$, the Bayesian agent computes the posterior distribution as
\begin{equation}
p(x|s) = \frac{\mu(x)\alpha(s|x)}{\sum\limits_{y\in\mathcal{X}}\mu(y)\alpha(s|y)}.
\label{eqn:stateposterior}
\end{equation}
The  agent has utility function $u(x,a)$ over the states $x\in\mathcal{X}$ and actions $a\in\mathcal{A}$.
\begin{definition}
  An agent satisfies attention rationality if there exists an information acquisition cost $C(\mu,\alpha)\in\mathbb{R}^{+}$ such that for the selected attention function $\alpha$, it selects actions $a\in\mathcal{A}$ 
  %and attention function $\alpha(s|x)$ 
  that satisfy the following conditions:
%  (where $\E$ denotes the expectation operator):
\begin{compactenum}[i)]
\item Expected Utility Maximization:
\begin{equation}
a^* \in\argmax_{a \in \mathcal{A}} \E\{ u(x,a) | s\} = \argmax_{a \in \mathcal{A}} \left\{\sum\limits_{x\in\mathcal{X}}p(x|s)u(x,a)\right\} \quad \forall p(x|s)\in\mathcal{S}(\alpha)
\label{eqn:utilitymaximization}
\end{equation}
\item Attention Selection Rationality: 
\begin{align}
&\alpha^*(s|x) \in\argmax_{\alpha\in \boldsymbol{\alpha}} \Big\{\E_{s\in\mathcal{S}(\alpha)}\{\operatorname*{max}_{a\in\mathcal{A}}[\sum_{x\in\mathcal{X}}p(x|s)u(x,a)]\}-
 C(\mu,\alpha)\Big\}
\label{eqn:attentionmaximization}
\end{align}
where $C(\mu,\alpha)$ is the information acquisition cost of attention function $\alpha$ for the prior distribution $\mu$. Also, $\boldsymbol{\alpha}$ denotes the $\mathcal{S}(\alpha)-1$ dimensional unit simplex of probability mass functions. 
\end{compactenum}
\label{def:rationalinattention}
\end{definition}
\noindent
%\textcolor{blue}{In~(\ref{eqn:attentionmaximization}),}
Eq.\ (\ref{eqn:utilitymaximization}) states that the agent  selects actions that are consistent with Bayesian utility maximization. In~(\ref{eqn:attentionmaximization}), the probability mass function $\alpha^{\ast}(s|x)$ is the optimal attention function selected by the agent  to maximize the expected utility.

The information acquisition cost $C(\mu,\alpha)$ will be interpreted in the convex case in Sec.\ \ref{sec:MutualInformationMeasuresandAttentionCost} as a behavioral economics constraint (see~(18)). Then~(\ref{eqn:attentionmaximization}) can be viewed as maximizing the Lagrangian of the following information acquisition cost constrained Bayesian utility maximization problem
\begin{equation*}
\begin{split}
\alpha^*(s|x) \in & \argmax_{\alpha\in \boldsymbol{\alpha}} \Big\{\E_{s\in\mathcal{S}(\alpha)}\{\operatorname*{max}_{a\in\mathcal{A}}[\sum_{x\in\mathcal{X}}p(x|s)u(x,a)]\}\Big\}\\
& \text{s.t. } C(\mu,\alpha) \leq K, \quad K\in\mathbb{R}^{+}
\end{split}
\end{equation*}
where the Lagrange multiplier of the constraint is set to $1$.
%\textcolor{blue}{another interpretation of the methodology used in the paper is that of inverse reinforcement learning of a contextual bandit,/ a contextual bandit is a ...... analog map/ In our case, the context is behavioral economics based and hasnt been solved.  add in abstract, intro, viewpoint 2, conclusion.}
%\textcolor{blue}{The rationally inattentive Bayesian setting closely resembles constrained contextual bandits~(\cite{BD14}) wherein the agent at each iteration receives a reward drawn from a distribution depending on the action and a context vector. In addition, there is an associated cost for each action choice.  The agent seeks to maximize its cumulative expected reward less the cost incurred for its action choices. In the formulation explained above, the action taken by the agent is analogous to the attention function $\alpha(s|x)$ (\ref{eqn:attentionmaximization}), the cost $C(\mu,\alpha)$ corresponds to the action cost and the decision problem is equivalent to the current context. The expected reward for choosing action $\alpha(s|x)$ under context $k$ (decision problem) is given by the objective function in~(\ref{eqn:attentionmaximization}).  However, the two models of rational inattention and constrained contextual bandits differ in the following way- the agent in the latter case has no knowledge of the expected reward as a function of the action and context whereas in the former case, the agent precisely knows the expected reward if the agent is rational inattentive. ADD- We are doing inverse contextual bandits???}

\subsubsection*{Viewpoint 2. Inverse Reinforcement learning: Observer's Model and Deep Clustering of Frames}
In inverse reinforcement learning, an observer (analyst)  seeks to estimate the utility function of a decision maker by observing its actions in response to a stimulus input~(\cite{NG00}\footnote{\cite{NG00} can be viewed as a special case of the NIAS condition (\ref{eqn:MILPagentutility}) defined in Sec.\ \ref{sec:DecisionTestforRationalInattentionandRecoverabilityofAgentPreferences}. Specifically they consider an infinite horizon discounted cost Markov decision process. Then given the stationary policy and transition probabilities, it is straightforward to construct a set of inequalities that the cost function satisfies. Indeed inverse optimal control dates back to \cite{KLM64}. The formulation in Sec.\ \ref{sec:DecisionTestforRationalInattentionandRecoverabilityofAgentPreferences} (and in \cite{CD15}) is more general since the framework involves a Bayesian agent (partially observed system where the analyst does not have access to the observations or observation probabilities of the agent) with the added complexity of rational inattention constraints. Despite this added generality, the NIAS~(\ref{eqn:bayesianexpectedutility}) and NIAC~(\ref{eqn:attentionmaximizationthrm}) conditions (Theorem~\ref{thrm:decisiontestrationalinattention}) are necessary and sufficient for detecting a constrained Bayesian utility maximizer.}).
The revealed preference framework we consider is more general: by observing the actions of the agent,  the observer (analyst) first aims to determine
if the agent is rationally inattentive, and if so, estimate the agent's utility function and information acquisition  cost.  
The \textit{action selection policy} of the agent $(\pi(a|x,f))$ is the conditional probability of choosing an action $a\in \mathcal{A}$ given state $x\in \mathcal{X}$ and is defined as:
\begin{equation}
\pi(a|x,f) = \sum_{s\in\mathcal{S}(\alpha)} \eta(a|s)\alpha(s|x),
\label{eqn:ActionSelection}
\end{equation}
where $\eta(a|s)$ is called the \textit{choice function} i.e.\ the probability of choosing an action given the signal $s\in\mathcal{S}(\alpha)$. Note that in (\ref{eqn:ActionSelection}), the probability of choosing an action depends only on the signal realization $s\in\mathcal{S}(\alpha)$ and not only the true state $x$ for a rationally inattentive agent. Thus, the choice function $\eta(a|s)$ in~(\ref{eqn:ActionSelection}) ensures the data is matched i.e.\ unobserved attention function $\alpha(s|x)$ is consistent with the observed action selection policy. The observer (analyst) has access to the  dataset of states $x_t$ and actions $a_t$ chosen by the agent for videos indexed by 
$t=\{1,\ldots, T\}$, where $T=140,000$ videos~(as discussed in Sec.\ \ref{sec:RationalInattentionandUtilityMaximizationintheYouTubeSocialNetwork}):
\begin{equation}
\mathcal{D} = \{(x_t,f_t,a_t)\}_{t=1}^T.
\label{eqn:observationdataframe}
\end{equation}
 In~(\ref{eqn:observationdataframe}), the parameter $f_t$  represents all the framing information immediately apparent to the agent. Typically, framing information $f_t$  includes images, video, text, and  data. In our YouTube example, $f_t$ maps the title and thumbnail of a video to
an  integer representing a unique frame. Qualitatively, different values of $f_t$ determine different action policies  by the agent for a given title and thumbnail.
An important issue when applying rational inattention theory is accounting for the agent's framing effects that impact the agent's behavior. To account for framing effects, we assume there are $\{0,1,\dots,N\}$ possible  frames. In Sec.\ \ref{sec:FrameingInformationDeepLearning} a deep embedded clustering method is used to construct $f_t$ given the title and thumbnail of the YouTube video indexed by $t=\{1,2\dots T\}$, where $T=140,000$ videos~(as detailed in Sec.\ \ref{sec:RationalInattentionandUtilityMaximizationintheYouTubeSocialNetwork}). 

Given the set of frames, rational inattention theory aims to determine if the dataset $\mathcal{D}$ is consistent with Definition~\ref{def:rationalinattention}. To test for rational inattention we require  estimates of the (possibly randomized)  action selection policy $\pi(a|x,f)$ (\ref{eqn:ActionSelection}) and prior beliefs $\mu(x)$ of the agent. Using dataset $\mathcal{D}$, the maximum likelihood estimates of $\pi(a|x,f)$ and $\mu(x)$ are 
\begin{equation}
\hat{\pi}(a|x,f) = \frac{\sum_{t=1}^T\mathbf{1}\{x_t=x,a_t=a, f_t=f\}}{\sum_{t=1}^{T}\mathbf{1}\{x_t=x, f_t = f\}}, \quad
\hat{\mu}(x) = \frac{1}{T}\sum_{t=1}^T\mathbf{1}\{x_t=x\},
\label{eqn:policyprior}
\end{equation}
where $\mathbf{1}\{\cdot\}$ is the indicator function. Given the maximum likelihood estimates (\ref{eqn:policyprior}), Sec.\ \ref{sec:DecisionTestforRationalInattentionandRecoverabilityofAgentPreferences}  provides a decision test for rational inattention; Sec.\ \ref{sec:MutualInformationMeasuresandAttentionCost} specifies a test for utility maximization with R{\'e}nyi mutual information based rational inattention cost of information acquisition. For agents that satisfy the rational inattention test in Theorem~\ref{thrm:decisiontestrationalinattention}, methods to recover their utility function $u(x,a,f)$, attention function $\alpha(s|x)$, posterior distribution $s(x)$, and information acquisition cost $C(\mu,\alpha)$ are also provided in Sec.\ \ref{sec:DecisionTestforRationalInattentionandRecoverabilityofAgentPreferences}. 

In Appendix~\ref{appdx:Risk-AwareUtilityMaximizationbyOff-PolicyEstimation}, we present a finite sample analysis of the agent's action selection policy. 

\subsection*{Constrained Contextual Bandits}
The rationally inattentive Bayesian model of the agent described in Viewpoint~$1$ above can be interpreted as a constrained contextual bandit~(\cite{BD14}). In contextual bandits, the agent receives a reward drawn from a distribution depending on the action and a context vector. In addition, there is an associated cost for each action choice. The agent seeks to maximize its cumulative expected reward less the cost incurred for its action choices. In our YouTube case, the arm pulled by the agent is equivalent to the attention function $\alpha(s|x)$ chosen by the agent (i.e. we assume a continuum of arms), the cost of pulling an arm becomes the behavioral economics-based rational inattention cost $C(\mu,\alpha)$ and the context corresponds to decision problems defined in  Sec.\ \ref{sec:RationalInattentionandUtilityMaximizationintheYouTubeSocialNetwork}. The expected reward for choosing action $\alpha(s|x)$ given context $k$ (decision problem) is the objective function in~(\ref{eqn:attentionmaximization}) which is maximized by the agent. Having a continuum of arms is more general than a finite arm bandit problem. Indeed, the  conditions in Theorem~\ref{thrm:decisiontestrationalinattention} still remain necessary and sufficient for utility maximization for a discrete set of admissible attention functions (arms).

Viewpoint~$2$ corresponds to inverse reinforcement learning of a contextual bandit with rational inattention constraints. As the observer (analyst), our aim is to  decide if the agent is a constrained Bayesian utility maximizer and if so,  reconstruct the utility function of the agent by observing its actions. Thus, as observers (analysts), we do inverse reinforcement learning of a constrained contextual bandits problem. To the best of our knowledge, inverse reinforcement learning for constrained contextual bandits has not been addressed in the literature.
%%%%%%%%%%%%%%%%%%%%%%%%%%%%%%%%%%%%%%%%%%%%%%%%%%%%%%%%%%%%%%%%%%%%%%%%%%%%%%%%%%%%

\section{Constructing Preference and Policy Invariant Frames via Deep Embedded Clustering}
\label{sec:FrameingInformationDeepLearning}
Recall from Sec.\ \ref{sec:Introduction} that in behavioral economics, "framing"  describes information an agent has when making a decision.
Framing information can dramatically impact the agent's action selection policies in YouTube. Specifically, the title and thumbnail have a significant impact on the agent's commenting behavior as reported in~(\cite{HAK17}). Such framing effects must be accounted for to minimize the probability of incorrectly rejecting a rationally inattentive agent. To account for these framing effects a deep embedding method is provided that learns the policy invariant frames of the agent. Specifically, a mapping of $f_t$ to $n_t\in\{1,\dots,N\}$ is constructed where for each $n\in\{1,\dots,N\}$ the behavior of the agent is invariant. In the YouTube social network the framing information available to the agent is comprised of the title and thumbnail of each video. Given that agents are ordinal preference invariant to minor variations in the title and thumbnail, it is possible to map the features $f_t$ to one of $\{1,\dots,N\}$ discrete frames learned using deep embedding. 

\begin{figure}[h!]
  \centering
\begin{tikzpicture}[font = \normalsize, scale =0.8,transform shape]
% Define block styles
\tikzstyle{block} = [draw,black,rectangle,fill= white,align=center]

%Constants
\def\xo{1.0}
\def\x{3.1}

%Nodes
\node (FI) at (-2,0) {$f_t$};
\node (FO) at (8.5,0) {$\hat{f}_t$};
\node (KO) at (8.5,-2+0.5) {$n_t$};
\node (KI) at (4,-2.25+0.5) {$N$};

\node[block] (PK) at (0,0) {Prior Knowledge \\ $w(f)$};
\node[block] (NS) at (2.35,1.25) {Noise};
\node[draw,circle,inner sep=0pt,minimum size=3pt] (N) at (2.35,0) {\Large $+$};
\node[block] (E) at (4,0) {Encoder \\ $r(\tilde{f})$};
\node[block] (D) at (7,0) {Decoder \\ $g(z)$};
\node[draw,fill=black,circle,inner sep=0pt,minimum size=2pt] (LS) at (5.5,0) {\_};
\node[block] (CL) at (7,-2+0.5) {Frame \\ Association};

% Paths
\draw[black, ->] (FI) -- (PK);
\draw[black, ->] (PK) -- (N);
\draw[black, ->] (N) -- (E);
\draw[black, ->] (E) --node[midway,above]{$z_t$} (D);
\draw[black, ->] (NS) --node[midway,right]{$\varepsilon$} (N);
\draw[black, ->] (D) -- (FO);
\draw[black, ->] (LS) |- (CL);
\draw[black, ->] (CL) -- (KO);
\draw[black, ->] (KI.east) -- (KI-|CL.west);
\end{tikzpicture}
    \caption{Schematic of the deep embedded clustering method to map the framing information $f_t$ to the discrete frame $\{1,\dots,N\}$. The parameter $w(f)$ contains all prior knowledge of the input framing information, $\varepsilon$ is a Gaussian white noise term, $r(\tilde{f})$ is the encoder, $z_t$ is the latent space representation of $f_t$, $g(z)$ represents the decoder, and $\hat{f}_t$ is the output of the autoencoder.}
\label{fig:deepembeddedclustering}
%\vspace{-5pt}
\end{figure}
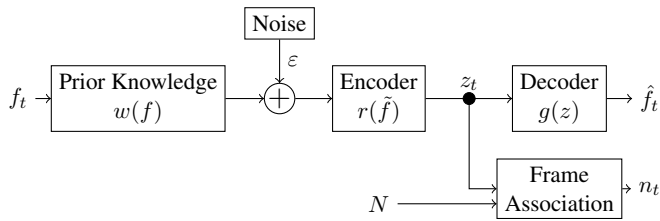

The deep embedding method uses natural language processing and image processing tools in an autoencoder (see Appendix~\ref{appdx:DeepClustering}) to construct the latent representation $z_t$ of $f_t$, and includes a clustering layer to simultaneously learn how to associate each $f_t$ to one of $\{1,\dots,N\}$ discrete frames. A schematic of the clustering method is illustrated in Fig.~\ref{fig:deepembeddedclustering}. 

The autoencoder comprises two deep neural networks, the first is the encoder that  maps the input $f_t$ to the latent space representation $z_t$, and the second is the decoder that map the latent space representation $z_t$ to the input $f_t$ where $\hat{f}_t\approx f_t$. To force the encoder to learn robust latent representations, the autoencoder is trained using corrupted versions of the input. Such an autoencoder is known as a denoising autoencoder~(\cite{VLBM08,Ben09}). The denoising autoencoder  encodes the input into the latent space representation, and attempts to remove the effect of the corruption process stochastically applied to the input of the autoencoder. Removing effects of the corruption process is performed by learning the statistical dependencies between the inputs. A detailed description of the denoising autoencoder architecture is in Appendix~\ref{appdx:DeepClustering}  with focus on the title and thumbnail of YouTube videos. 

Though the latent space representation of the input has been used extensively for clustering, such methods are not guaranteed to preserve any intrinsic local structure of the framing data $f_t$. To ensure the autoencoder both minimizes the reconstruction error and maximizes the intrinsic local structure of the data, a clustering loss is used. The loss of the deep embedded clustering method (Fig.~\ref{fig:deepembeddedclustering}) is:
\begin{equation}
L = ||s-g(f(w(s)+\varepsilon))||_2^2+\operatorname{KL}(P||Q)
\label{eqn:lossfunction}
\end{equation}
The first term in~(\ref{eqn:lossfunction}) is the reconstruction error, and $\operatorname{KL}(P||Q)$ denotes the Kullback-Leibler (KL) divergence  of the discrete probability distributions $P$ and $Q$. %The KL-divergence encodes the information gained if the probability distribution $P$ was used in place of the prior probability distribution $Q$.
Here $Q$ (with elements $q_{t,n}$) is the prior probability distribution of cluster association between the latent variables $z_t$ and the associated frames $n_t$. If we assume each cluster is generated from a Gaussian normal distribution with mean $\Psi_n$, then the normalized probability of association of each $z_t$ with every cluster head $\Psi_n$ is given by:
 \begin{equation}
q_{t,n} = \frac{1}{T}\left[ \frac{(1+||z_t-\Psi_n||^2)^{-1})}{\sum_{n=1}^N(1+||z_t-\Psi_n||^2)^{-1}}\right] \quad \forall n\in\{1,\dots,N\}.
\label{eqn:clusterprior}
\end{equation}
%Using the Student-t distributed similarity kernel~(\cite{GGLY17}), we first construct the symmetric probability distribution $Q$ pertaining to the association of each data point to the cluster head. We want to enhance the high confidence associations and weaken the low confidence associations, while also ensuring there is no degenerate clustering. We then construct a target probability distribution $P$ and the KL divergence loss ensures the autoencoder produces the desired effect.
Note that the above definition ensures $\sum_{t=1}^{T}\sum_{n=1}^{N} q_{t,n} = 1$. Given $Q$, to avoid degenerate clustering solutions which allocate most of the frames to a few clusters or assign a cluster to a sample outlier, the target distribution $P$ is designed with elements $p_{t,n}$ defined as: 
\begin{equation}
p_{t,n} = \frac{1}{T}\left[\frac{q_{t,n}^2/\sum_{t=1}^Tq_{t,n}}{\sum_{n=1}^N(q_{t,n}^2/\sum_{t=1}^Tq_{t,n})} \right],  \quad P(z_t=n)=\sum_{t=1}^Tq_{t,n} , \quad t = 1,\ldots, T
\label{eqn:clustertargetdist}
\end{equation}
Note that $P(z_t = n)$ is the empirical frequency that  $z_t$ belongs to cluster $n$.

From (\ref{eqn:clusterprior}) and (\ref{eqn:clustertargetdist}), if all the data-points are associated with a specific cluster this will increase the loss (\ref{eqn:lossfunction}). Additionally, if the cluster is associated with several data points with low-confidence, this will also increase the loss (\ref{eqn:lossfunction}). Minimizing the loss (\ref{eqn:lossfunction}) can be interpreted as a form of self-training as $P$ depends on $Q$. Specifically, in self-training we take an initial classifier and an unlabeled dataset, then label the dataset with the classifier in order to train on its own high confidence predictions. This ensures that the latent clusters are constructed to avoid outliers. 

The deep embedding method that maps $f_t$ to $n_t\in\{1,\dots,N\}$ is formalized in Algorithm~\ref{alg:framingassociation}; see Algorithm~\ref{alg:framingassociation_1} in Appendix~\ref{appdx:DeepClustering} for the detailed version. The pre-training step is used to initialize the encoder and decoder parameters prior to performing any clustering. This is a critical step as the initial latent space representation of $\{f_t\}_{t=1}^T$ is used to select the approximate locations of the $N$ latent space cluster centers $\Psi^o$. Given the pre-trained denoising autoencoder weights, we use the Lloyd heuristic algorithm to select the locations of the $N$ latent space cluster centers $\Psi^o$. Given the cluster centers, the deep clustering method is applied to minimize the loss (\ref{eqn:lossfunction}) by simultaneously adjusting the cluster associations and autoencoder weights. Note that in Algorithm~\ref{alg:framingassociation_1}, since the distribution $P$ (\ref{eqn:clustertargetdist}) depends on the weights of the encoder, we update $P$ after $\zeta$ iterations. This reduces the probability of  instability associated with cycling between adjusting weights and cluster associations. The final result of Algorithm~\ref{alg:framingassociation_1} is achieved when the change in cluster associations is below a  threshold  $\delta$. 
%To ensure only frames $f_t$ that can be confidently associated to one invariant frame, all frames that fail to satisfy $\operatorname{max}\{q_{t,n}\} \leq \delta_c$ are discarded. 

\makeatletter
\newcommand{\HEADER}[1]{\ALC@it\underline{\textsc{#1}}\begin{ALC@g}}
\newcommand{\ENDHEADER}{\end{ALC@g}}
\makeatother

\begin{algorithm} % enter the algorithm environment
\caption{Deep Embedded Clustering for Framing Association} % give the algorithm a caption
\label{alg:framingassociation} % and a label for \ref{} commands later in the document
\begin{algorithmic} % enter the algorithmic environment
    \REQUIRE Set of framing information $\{f_t\}_{t=1}^T$, number of unique frames $N$, stopping threshold $\delta\in(0,1)$, confidence threshold $\delta_c\in(0,1)$, and updating interval $\zeta$. 
    \HEADER{Pre-train}
    	\STATE Pre-train the denoising autoencoder without any frame association.
    \ENDHEADER
    \HEADER{Initialize}
    	\STATE Initialize the $N$ cluster centers $\Psi^o$ using k-means clustering in the latent space and set $\varepsilon=0$.
    \ENDHEADER
        \HEADER{Deep Clustering}
        \STATE Train the deep clustering autoencoder and frame association layers (refer to Appendix~\ref{appdx:DeepClustering}). 
    \ENDHEADER
\RETURN Invariant frames $n_t \forall t\in\{1,\dots,T\}$ such that $\operatorname{max}_n\{q_{t,n}\} > \delta_c$. 
\end{algorithmic}
\end{algorithm}

Given the preference and policy invariant frames $\{n_t\}_{t=1}^T$, we substitute $n_t\rightarrow f_t$ in $\mathcal{D}$ (\ref{eqn:observationdataframe}). Within each frame $f\in\{1,2\dots N\}$, we further divide the group of videos into $K$ sub-partitions or \textit{decision problems}. The procedure for further partitioning each frame is explained in detail in Sec.\ \ref{sec:YTParameters}.
Using $\mathcal{D}$ with the invariant frames and their corresponding decision problems, Sec.\ \ref{sec:DecisionTestforRationalInattentionandRecoverabilityofAgentPreferences} and Sec.\ \ref{sec:MutualInformationMeasuresandAttentionCost} illustrate how to detect if the agent is rationally inattentive for different information acquisition cost constraints, and how to recover the utility functions.

\section{Decision Test for Rational Inattention; Estimating Utility/ Attention Costs}
\label{sec:DecisionTestforRationalInattentionandRecoverabilityofAgentPreferences}

Here we construct a decision test for rational inattention (Definition~\ref{def:rationalinattention}). The resulting preference-based inverse reinforcement learning algorithm uses  the observed stochastic choice dataset $\mathcal{D}$ (\ref{eqn:observationdataframe}) and invariant frames $f\in\{1,2\dots M\}$. Recall that the group of YouTube videos in each frame is further sub-divided into $K$ decision problems and their associated utility functions $\{u_k(x,a)\}_{k=1}^K$ as defined in~(\ref{eqn:utilitymaximization}). Theorem~\ref{thrm:decisiontestrationalinattention} below is our main result and is a slight generalization of \cite{CD15} and \cite{CM15}. The proof is in Appendix~\ref{appdx:proofs}.
\begin{theorem}
The dataset $\mathcal{D}$ (\ref{eqn:observationdataframe}) satisfies rational inattention (Definition~\ref{def:rationalinattention}) iff  the action selection policy~$\pi_k(a|x,f)$ defined in (\ref{eqn:ActionSelection}) and utility function $u_k(x,a,f)$ (\ref{eqn:utilitymaximization}) for decision problem $k \in \{1,2\dots K\}$ in frame $f\in\{1,2\dots N\}$  satisfy the following two conditions:
%\begin{equation}
%\pi_k(a|x,f) = \sum_{s\in\mathcal{S}(\alpha_k)}\alpha_k(s|x,f)\eta_k(a|s,f), \quad
%\mathcal{S}(\alpha_k) = \{p_k(x|a,f): a\in\mathcal{A}_k\} \nonumber
%\label{eqn:datamatching}
%\end{equation}
%where the choice function  $\eta_k(a|s)$ is the probability of selecting action $a$ given the posterior  associated with signal $s\in\mathcal{S}(\alpha_k)$. 
%\textcolor{blue}{Additionally, the following two conditions must be satisfied.} 
%\begin{compactenum}[i)]
%\item The utility  functions $\{u_k(x,a,f)\}_{k=1}^K$ ($k$ indexes the decision problem) satisfy the following inequalities for each $f\in \{1,2\dots T\}$
\begin{align}
(1) \textbf{ NIAS: }&\sum_{x\in\mathcal{X}}p_k(x|a,f)[u_k(x,a,f)-u_k(x,b,f)] \geq 0 \quad \forall a,b\in\mathcal{A}_k \nonumber\\
&p_k(x|a,f) = \frac{\mu(x)\pi_k(a|x,f)}{\sum_{y\in\mathcal{X}}\mu(y)\pi_k(a|y,f)}
\label{eqn:bayesianexpectedutility}\\
%\end{align}
 %Also, for any sequence of decision problems $\mathcal{A}_1,\mathcal{A}_2\dots %\mathcal{A}_L (L\leq K)$, the action selection policies $\{\pi_k(a|x,f)\}_{k=1}^L$ satisfy
%\begin{align}
(2) \textbf{ NIAC: }&\sum_{k=k_1}^{k_L}G_{k,k}-G_{k+1,k} \geq 0 \quad \text{for any sequence of decision problems } k_1,k_2\dots k_L (L\leq K) \label{eqn:attentionmaximizationthrm} \\
&G_{k,w} = \sum_{s\in\mathcal{S}(\alpha_k)}\sum_{x\in\mathcal{X}}\mu(x)\pi_k(a|x,f)\operatorname*{max}_{b\in\mathcal{A}_w}\left\{\sum_{x\in\mathcal{X}}p_k(x|a,f)u_w(x,b,f)\right\} \quad (k_{L+1}=k_1) \nonumber
  %\alpha_k(s|x,f) =\sum_{a\in\mathcal{A}_k}\pi_k(a|x,f)\mathbf{1}\{p_k(x|a,f) = s\}, 
  %&\text{ with $$. } \nonumber
\end{align}
%\end{compactenum}
\label{thrm:decisiontestrationalinattention}
\end{theorem}
Theorem \ref{thrm:decisiontestrationalinattention} is proved in Appendix~\ref{appdx:proofs}. It specifies necessary and sufficient conditions for an agent to be a  Bayesian utility maximizer~(\cite{CD15}), namely, \textit{No Improving Attention Switches} (NIAS) and \textit{No Improving Attention Cycles} (NIAC). 

Theorem~\ref{thrm:decisiontestrationalinattention} is best viewed from the point of view of an observer (analyst) performing inverse reinforcement learning on a Bayesian agent. If~(\ref{eqn:bayesianexpectedutility}) and (\ref{eqn:attentionmaximizationthrm}) have no feasible solutions, then the agent does not satisfy rational inattention based Bayesian utility maximization.  According to Theorem~$\ref{thrm:decisiontestrationalinattention}$, the analyst only needs to know the action selection policy $\pi_k(x|a,f)$ defined in~(\ref{eqn:ActionSelection}); this can be estimated from the dataset as described in (\ref{eqn:policyprior}). Note that the analyst does not require knowledge of the agent's attention function $\alpha_k(s|x,f)$ defined in (\ref{eqn:attentionmaximization}) even though this attention function together with the utility $u_k(\cdot)$ determines the actions of the $k^{th}$ agent.

A few words about NIAS and NIAC. The NIAS condition implies that the agent's actions are optimal under posterior beliefs. The NIAC condition implies that the sum of expected utilities~(\ref{eqn:attentionmaximization}) over decision problems cannot be increased by reassigning attention selection policies.
For readers familiar with revealed preference theory, the NIAC condition is analogous to the GARP conditions in Afriat’s theorem~(\cite{Var12,Die12}) for testing utility maximization behavior. $G_{k,w}$ in~(\ref{eqn:attentionmaximizationthrm}) gives the expected utility of using attention function $\alpha_k(s|x,f)$.  Additionally, one can construct the total expected ordinal utility of an agent given by 
\begin{equation}
V(\boldsymbol{\pi}(a|x,f)) = \sum_{k=1}^K\sum_{x\in\mathcal{X}}\sum_{a\in\mathcal{A}_k}\pi_k(a|x,f)\mu(x)u(a,x,f).
\label{eqn:totalexpectedutility}
\end{equation}
As an aside, in Appendix~\ref{appdx:Risk-AwareUtilityMaximizationbyOff-PolicyEstimation} we illustrate how the observer (analyst) performing inverse  reinforcement learning  can maximize  $V(\boldsymbol{\pi})$ w.r.t policies $\boldsymbol{\pi}$ so as to provide optimal behavioral recommendations to agents. 
% Utility functions for agents can be constructed iff NIAC and NIAS hold.  
%Eq.\ (\ref{eqn:datamatching}) in Theorem~\ref{thrm:decisiontestrationalinattention} ensures the data in our dataset $\mathcal{D}$ is matched, i.e, the attention function $\alpha_k(s|x,f)$ and action selection policy $\eta_k(a|s,f)$ result in the observed action selection policy $\pi_k(a|x,f)$ defined in (\ref{eqn:ActionSelection}). 
%The inequalities (\ref{eqn:bayesianexpectedutility}), (\ref{eqn:attentionmaximizationthrm}) together ensure that the agent satisfies Bayesian expected utility maximization.

From a practical point of view, (\ref{eqn:bayesianexpectedutility}), (\ref{eqn:attentionmaximizationthrm}) in Theorem~\ref{thrm:decisiontestrationalinattention} can be implemented by solving an equivalent mixed-integer linear feasibility problem as described below in Corollary~\ref{corl:MILP}.
 %can be equivalently formulated as a mixed integer linear program (MILP) as shown in the corollary below
 %\vspace{0.5cm}
\begin{corollary} \label{corl:MILP}
The NIAS  and NIAC conditions in Theorem~\ref{thrm:decisiontestrationalinattention} can be equivalently expressed as the following mixed-integer linear feasibility problem:\newline
% Utility functions $\{u_k(x,a,f)\}_{k=1}^K$ ($k$ indexes the decision problem) exist that satisfies the constraints
%\begin{equation*}
 %\mathcal{L}( \{u_k(x,a,f)\}_{k=1}^K ) \quad \forall f\in\{1,2\dots T\}  
 %\quad\text{ for }  f \in \{1,2,\ldots, N\}
 %\label{eqn:MILPAgentutility}
 %\end{equation*}: 
 Find $u_k(x,a,f)\in[0,1], ~\delta_{b,k,f}, \zeta_{b,k,f} \in\{0,1\}~ ~\forall a,b\in\mathcal{A}_k,~k\in\{1,2\dots K\},f\in\{1,2\dots N\}$ such that
 \begin{equation} \label{eqn:MILPagentutility}
 \begin{split}
 \textbf{NIAS}:\quad&\sum_{x\in\mathcal{X}}p_k(x|a,f)[u_k(x,a,f)-u_k(x,b,f)] \geq 0\\
 \textbf{NIAC}:\quad&\sum_{k=k_1}^{k_L}\left(\sum_{a\in\mathcal{A}_k}p_k(a|f)m_{k}(a)-\sum_{a\in\mathcal{A}_{k+1}}p_{k+1}(a|f)n_{k+1}(a)\right) \geq 0,  \quad \forall k_{1:L}\in\{1,\dots K\}
 \end{split}
 \end{equation}
 \begin{equation} \label{eqn:bigM}
 \begin{split}
 &m_k(a,b) = \sum_{x\in\mathcal{X}}p_k(x|a,f)u_k(x,b,f) \\
 &m_k(a) \geq m_k(a,b) \quad \forall a,b\in\mathcal{A}_k \\
 &m_k(a) \leq m_k(a,b) + M(1-\delta_{b,k,f}), \quad \sum_{b\in\mathcal{A}_k}\delta_{b,k} = 1 \\
 &n_{k+1}(a) \geq m_{k+1}(a,b) \quad \forall a\in\mathcal{A}_{k+1}, \quad \forall b\in\mathcal{A}_k \\
 &n_{k+1}(a) \leq m_{k+1}(a,b) + M(1-\zeta_{b,k,f}), \quad \sum_{b\in\mathcal{A}_k}\zeta_{b,k,f} = 1
 %&u_k(x,a)\in[0,1], \quad \delta_{b,,fk}, \zeta_{b,k} \in\{0,1\} \\
 %&\forall a,b\in\mathcal{A}_k, \quad \forall k\in\{1,2,\dots,K\} 
 \end{split}
 %\label{ineq:Feasibility}
 \end{equation}
 where $k_{L+1}=k_1,~L\leq K$ and $M \in \mathbb{R}^{+}$ is a large positive real constant.
 %\label{ineq:MILP}
 \end{corollary}
%\end{compactenum}
%\label{thrm:decisiontestrationalinattention}
%\end{theorem}
%In (\ref{eqn:MILPagentutility}), $M$ is used
%we use $M$(a large constant) in the context of the BIg M Method() to convert a non-linear problem of finding the maximum to a set of linear inequalities using binary valued integer variables. 
%Note that statement Theorem~\ref{thrm:decisiontestrationalinattention}~($ii$) is an equivalent formulation for~(\ref{eqn:bayesianexpectedutility}) and (\ref{eqn:attentionmaximizationthrm}) in Theorem~\ref{thrm:decisiontestrationalinattention}. 
%and binary valued integer variables to convert the non-linear problem of finding a maximum over a set to a system of linear inequalities, a technique similar to the Big M method~(\cite{BigM}). 
%To determine if a $u(x,a,f)$ exists for the constraint set can be evaluated using a variety of numerical methods including branch-and-bound, cutting planes, branch-and-cut, and branch-and-price~(\cite{GG11}).
Corollary~\ref{corl:MILP} provides a constructive feasibility test for the analyst to determine if YouTube user groups in dataset $\mathcal{D}$ satisfy rational inattention based Bayesian utility maximization. It also provides a set-valued estimate of the user groups' utility function $u_k(x,a,f)$ that rationalizes the dataset $\mathcal{D}$.
%a deterministic test to check if utility functions can be constructed that rationalizes the given data. 
%In other words, the feasibility of the MILP in condition~($\ref{ineq:MILP}$) of Theorem~\ref{thrm:decisiontestrationalinattention} implies that the available data may be generated by an agent employing an arbitrary cognitive information cost structure to optimally take decisions in a bayesian sense under any of the considered decision problems. 
To justify our findings on the YouTube dataset, in Sec.\ \ref{sec:RationalInattentionandUtilityMaximizationintheYouTubeSocialNetwork} we perform robustness tests for every segment (groups of videos) in the YouTube dataset $\mathcal{D}$ (see Sec.\ \ref{sec:DeepClustering} and Sec.\ \ref{sec:CatDissect} for partitioning of the dataset into segments) to see how close they are to satisfying utility maximization behavior. 
%Corollary~\ref{corl:MILP} is the equivalent mixed integer linear program~(MILP) formulation of Theorem~\ref{thrm:decisiontestrationalinattention}. 

Eq.\  (\ref{eqn:bigM}) uses the Big M method~(\cite{BigM}) to express $G_{k,w}$ defined in~(\ref{eqn:attentionmaximizationthrm}) as a set of linear inequalities using binary valued integer variables.
%is expressed using we use $M \in \mathbb{R}_{+}$ (a large positive real constant) in the context of the Big M Method~(\cite{BigM}) to convert a maximization problem to a set of linear inequalities using binary valued integer variables. 
The resulting mixed-integer linear feasibility problem can be solved using a variety of numerical methods including branch-and-bound, cutting planes, branch-and-cut, and branch-and-price~(\cite{GG11}).

{\bf Construction of Information Acquisition Cost}: Given the utility function $\{u_k(x,a,f)\}_{k=1}^K$ that satisfy the constraints $(12)$ and $(13)$, the observer (analyst) can then construct an ordinal estimate of the associated cost of information acquisition $C(\mu,\alpha_k)$ of each attention function $\alpha_k$. Specifically, the ordinal cost of information acquisition $C(\mu,\alpha_k)$ can be computed by solving the following convex feasibility problem: 
\begin{align}
&G_{k,k}-G_{w,k} \geq C(\mu,\alpha_k)-C(\mu,\alpha_w) \nonumber\\
&C(\mu,\alpha_k) \geq 0 \quad \forall w,k\in\{1,\dots,K\}.
\label{eqn:ordinalcost}
\end{align}
Recall that if a solution to (12) exists, then a solution to (\ref{eqn:ordinalcost}) is guaranteed to exist from Theorem~1 and (3). Also, if the cost of a particular attention function is zero, then absolute bounds can be placed on the information acquisition cost of each attention function. For example if $C(\mu,\alpha_w)=0$, then the cost $C(\mu,\alpha_k) \in [G_{k,w}-G_{w,w}, G_{k,k}-G_{w,k}]$. The estimated cost function satisfies weak monotonicity in information--that is, if the attention function provides more information then it incurs a higher information acquisition cost. 
%However, it may be the case that the actual cost of information used by the agent does not satisfy this condition. In fact, only requiring rational inattention with no further restrictions on information cost does not impose any testable conditions for information monotonicity. 
%%%%%%%%%%%%%%%%%%%%%%%%%%%%%%%%%%%%%%%%%%%%%%%%%%%%%%%%%%%%%%%%%%%%%%%%%%%%%%%%%%

\section{R{\'{e}}nyi Entropy Information Acquisition Cost for Rational Inattention}
\label{sec:MutualInformationMeasuresandAttentionCost}
In this section we impose a behavioral economics structure to the information acquisition cost $C(\mu,\alpha)$ in (\ref{eqn:attentionmaximization}) which defines the attention function of a rationally inattentive agent. 
Sims' pioneering work~(\cite{Sim10}) uses Shannon mutual information between the prior distribution and attention function ($\alpha$) to define information acquisition cost in such agents whereas here the more general R{\'{e}}nyi mutual information is considered. The R{\'{e}}nyi mutual information between the prior $\mu(x)$ of the state and the selected attention function $\alpha_k(s|x)$ ($k$ denotes the $k^{th}$ decision problem (\ref{eqn:bayesianexpectedutility})) is
\begin{equation}
I_\beta(\mu,\alpha_k) =
   \begin{dcases}
     \frac{1}{\beta-1}\operatorname{ln}\left(\sum_{x\in\mathcal{X}}\sum_{a\in\mathcal{A}}\frac{p^\beta(x,a)}{\mu^{\beta-1}(x)p^{\beta-1}(a)}\right) \quad \beta\in(0,1)\cup(1,\infty)\\
     I(\mu,\alpha_k) \quad \beta = 1\\
     -\operatorname{ln}\left(\sum_{x\in\mathcal{X}}\sum_{a\in\mathcal{A}}\mu(x)p(a)\mathds{1}\{p(x,a) > 0\}\right) \quad \beta = 0
   \end{dcases}
   \label{eqn:renyimutualinformation}
\end{equation}   
where $\beta\in[0,\infty)$ is the R{\'{e}}nyi order. Note that we replace the signal "$s$" with the action "$a$" since we assume a one-to-one mapping between the two (parsimonious representation). The mutual information defined in (\ref{eqn:renyimutualinformation}) stems from the R{\'{e}}nyi entropy defined as
\begin{equation}
 H_{\beta}(X) = \frac{1}{1-\beta} \log\big(\ \sum_{i=1}^{n} p_{i}^{\beta} \big).
\end{equation} 
The R{\'{e}}nyi entropy is useful for measuring the information acquisition  cost since the parameter $\beta$ allows one to adjust the sensitivity of the cost to the shape of $\mu(x)$ and $\alpha_k(s|x)$. Indeed, for suitable choice of $\beta$ the R{\'{e}}nyi entropy includes the Hartley entropy, Shannon entropy, collision entropy and minimum entropy as special cases. 

An important feature of (\ref{eqn:renyimutualinformation}) is that for $\beta\in(0,1]$ the mutual information constraint is convex in $p(x,a)$ and $\mu(x)p(a)$. Also for $\beta > 1$, the information constraint is convex in $\mu(x)p(a)$ and quasi-convex in $p(x,a)$~(\cite{VH14,HV15,XE10}). In terms of (\ref{eqn:utilitymaximization}), the R{\'e}nyi mutual information cost constrained decision problem is
\begin{align}
&p_k^*(x,a) \in\argmax_{p(x,a)} \Big\{\sum_{a\in\mathcal{A}_k}\sum_{x\in\mathcal{X}}p(x,a)u(x,a)\Big\} \nonumber\\
&\quad\text{s.t.}\quad \mu(x) = \sum_{a\in\mathcal{A}_k}p(x,a) \quad \forall x\in\mathcal{X} \nonumber\\
  &\phantom{\quad\text{s.t.}\quad} I_\beta(\mu,\alpha_k) \leq \kappa_\text{max,k},
\quad p(x,a) \geq 0 \quad \forall x\in\mathcal{X}, a\in\mathcal{A}_k.
\label{eqn:shannoninformationmax}
\end{align}
In (\ref{eqn:shannoninformationmax}), $\kappa_\text{max}$ represents the maximum ``effort'' the agent is willing to invest to estimate the state $x\in\mathcal{X}$ prior to taking the action $a\in\mathcal{A}_k$ in decision problem $k\in\{1,\dots,K\}$. The constraint in~(\ref{eqn:shannoninformationmax}) constitutes the behavioral economics based constraint.

In addition to satisfying the NIAC and NIAS conditions in Theorem~\ref{thrm:decisiontestrationalinattention}, the information acquisition cost $C(\mu,\alpha)$ in Definition~\ref{def:rationalinattention} is now defined as:
\begin{equation}
C(\mu,\alpha) = \lambda I_{\beta}(\mu,\alpha) + \gamma,
\label{eqn:renyiform}
\end{equation} where $\lambda\in\mathbb{R}^{+},\gamma \in \mathbb{R}, \beta \in [0,1]$ and $I_{\beta}(\mu,\alpha_k)$ is as defined in~(\ref{eqn:renyimutualinformation}). Eq.\ (\ref{eqn:renyiform}) denotes the R{\'e}nyi Mutual Information cost and Shannon Mutual Information cost for $\beta \in (0,1)$ and $\beta=1$ respectively. In (\ref{eqn:renyiform}),  the objective function is linear and the constraint set is convex in the argument $p(x,a)$ for $\beta\in[0,1]$, hence necessary and sufficient conditions for the agent to satisfy rational inattention with the R{\'{e}}nyi/Shannon mutual information cost (\ref{eqn:renyiform}) can be constructed using the Karush-Kuhn-Tucker (KKT) conditions. Formally:
\begin{theorem}

%	\begin{compactitem}
Consider a Bayesian agent that satisfies the conditions~(\ref{eqn:bayesianexpectedutility}) and~(\ref{eqn:attentionmaximizationthrm}) in Theorem~\ref{thrm:decisiontestrationalinattention}. This agent satisfies utility maximization for
\begin{compactenum}[(i)]
\item  \textbf{R{\'e}nyi mutual information cost} (\ref{eqn:renyiform}) iff
% it satisfies the conditions in Theorem~\ref{thrm:decisiontestrationalinattention} and
%The information acquisition cost  for R{\'e}nyi/Shannon Mutual Information cost
%A rationally inattentive agent with utility function $u(x,a)$, joint-distribution $p(x,a)$, and $\beta\in(0,1)$ satisfies R{\'{e}}nyi mutual information cost (\ref{eqn:renyimutualinformation}) if and only if 
 there exist constants $\lambda_{1,k} > 0$ and $\lambda_{2,k}$  that satisfy the linear constraints 
%\textcolor{red}{(simplification of Eq.\ \ref{eqn:KKT_1})}
\begin{align}
&u_{k}(x,a) = \lambda_{1,k} \frac{\beta \eta^{\beta-1}( x, a)-(\beta-1)\mathbb{E}_{x}\left[\eta^{\beta-1}(x, a)\right] }{(\beta-1) {\left( \mathbb{E}\left[ \eta^{\beta-1}(x,a)\right] \right)}p^{\beta-1}( a)}     -\lambda_{2,k}, \nonumber\\
&\frac{1}{\beta-1}\operatorname{ln}\left(\E[\eta_{k}^{\beta-1}(x,a)]\right) = \kappa_\text{max}, \qquad
\eta_{k}(x,a) = \frac{p_{k}(x|a)}{p(x)},
\label{eqn:KKT_1}
\end{align}
for all $k\in\{1,2\dots K\}, x\in\mathcal{X}$, $a\in\mathcal{A}$.\\
	%for all $k\in\{1,2\dots K\}, x\in\mathcal{X}$, $a\in\mathcal{A}$.\\
% $\E[\cdot]$ is the expectation operator with respect to the joint distribution $p(x,a)$.
	\item \textbf{Shannon mutual information cost} (\ref{eqn:renyiform}) iff 
	%it satisfies the conditions in Theorem~\ref{thrm:decisiontestrationalinattention} and
%The information acquisition cost  for R{\'e}nyi/Shannon Mutual Information cost
%A rationally inattentive agent with utility function $u(x,a)$, joint-distribution $p(x,a)$, and $\beta\in(0,1)$ satisfies R{\'{e}}nyi mutual information cost (\ref{eqn:renyimutualinformation}) if and only if 
there exist constants $\lambda_{1,k} > 0$ and $\lambda_{2,k}$  that satisfy the linear constraints
\begin{align}
&u_{k}(x,a) = \lambda_{1,k}\operatorname{ln}(p_{k}(x|a))-\lambda_{2,k}, \quad \E\bigg[\operatorname{ln}\frac{p_{k}(x,a)}{p(x)p_{k}(a)}\bigg] = \kappa_\text{max}, 
%& \text{for all $k\in\{1,2\dots K\}, x\in\mathcal{X}$, $a\in\mathcal{A}$.}
\label{eqn:KKT_2}
\end{align}
for all $k\in\{1,2\dots K\}, x\in\mathcal{X}$, $a\in\mathcal{A}$.
\end{compactenum}
%\qed
%  $\E[\cdot]$ is the expectation operator with respect to the joint-distribution $p(x,a)$.
\label{thrm:reynicost}
\end{theorem}
%\begin{wrapfigure}{r
%	}{0.45\textwidth}
%	%\begin{figure}[h!]
%	\centering
%	\includegraphics[width=0.31\textwidth]{latent_2D_representation.jpg}
%	\vspace{0pt}
%%	\caption{t-SNE visualization of the latent space representation of title and thumbnail of YouTube videos constructed using Algorithm~\ref{alg:framingassociation} for the YouTube videos contained in the dataset $\mathcal{D}$.As seen, the t-SNE representation with $4$ frames (each indicated by a different color) indicates that the videos are sufficiently separated in the latent space.}
%	\label{fig:tsneunqiueframes}
%	%\end{figure}
%\end{wrapfigure}
%\begin{figure}
%	\begin{minipage}[t]{0.67\textwidth}
%			\includegraphics[width=0.31\textwidth]{latent_2D_representation.jpg}
%	\end{minipage}
%	\begin{minipage}[t]{0.3\textwidth}
%		\caption{
%			t-SNE visualization of the latent space representation of title and thumbnail of YouTube videos constructed using Algorithm~\ref{alg:framingassociation} for the YouTube videos contained in the dataset $\mathcal{D}$.As seen, the t-SNE representation with $4$ frames (each indicated by a different color) indicates that the videos are sufficiently separated in the latent space.
%		} \label{fig:tsneunqiueframes}
%	\end{minipage}
%\end{figure}
The proof is in Appendix~\ref{appdx:proof_KKT}. In Theorem~\ref{thrm:reynicost},  $\lambda_{1,k}>0$ and $\lambda_{2,k}$ are Lagrange multipliers pertaining to the inequality and  equality constraint in (\ref{eqn:shannoninformationmax}). Combining the linear equality constraints (\ref{eqn:KKT_1}), (\ref{eqn:KKT_2}) with the mixed integer linear program in Corollary~\ref{corl:MILP} yields a test for the R{\'{e}}nyi/ Shannon mutual information cost constrained optimization problem and  provides estimates of the associated  utility function of the agent. Thus we have constructed a preference based inverse reinforcement learning algorithm for the utility and information acquisition  cost of a Bayesian agent with a R{\'e}nyi and Shannon mutual information cost.

%\begin{comment}
%Theorem~\ref{thrm:reynicost}  ensures that the utility for different actions satisfies
%\begin{equation}
%u(x,a)-u(x,b) \propto \left(p(x|a)-p(x|b)\right)^{\beta-1} \quad\forall a,b\in\mathcal{A}.
%\end{equation}
%  For $\beta=1$, the utility difference between two actions changes, the relative probability of choosing those actions must change according to (\ref{eqn:constraintShannon}). Specifically, for $\beta = 1$, if the difference in utility increases, the difference in the probability of choosing the actions must also increase. However, for $\beta \in (0,1)$, then as the difference in utility increases, the difference in probability of choosing the two actions decreases. This illustrates the importance of how the parameter $\beta$ controls the behavior of the agent.
%  \end{comment}
  
  %%%%%%%%%%%%%%%%%%%%%%%%%%%%%%%%%%%%%%%%%%%%%%
%\section{Rational Inattention \&  Utility Maximization in YouTube Social Network}
\section{Rationally Inattentive Inverse Reinforcement Learning in YouTube Engagement}
\label{sec:RationalInattentionandUtilityMaximizationintheYouTubeSocialNetwork}

%Constructing utility based preference models for how users  interact and consume content in online social media platforms is important in social network analysis \cite{Kha17,HG17}. 

This section provides empirical evidence that the commenting behavior of YouTube users is consistent with Bayesian utility maximization with rational inattention. We consider a massive YouTube dataset comprising approximately $140000$ videos across $25,000$ channels and over $9$ millions users from April 2007 to May 2015. 

YouTube is an interesting example of a social network since the interaction between users includes video content. Users interact on  YouTube channels  by  posting comments and rating videos. In this paper, we analyze overall user engagement in YouTube videos. By inverse reinforcement learning in YouTube, we mean determining the existence and construction of utility functions in various user groups that rationalizes the commenting behavior in the YouTube dataset $\mathcal{D}$.
Extensive empirical  studies~(\cite{Kha17,HG17,ABCH15,HNK15,HAK17,AK17}) show  that the comments and ratings from users are  influenced by the thumbnail, title, category, and perceived popularity of each video. 
%As is typical in behavioral economics~(\cite{WK17}), by user behavior, we mean the average commenting behavior per YouTube channel, averaged over all the channels.

Before proceeding with details of our analysis, we briefly summarize our main conclusions:
\begin{compactenum}
\item Rationally inattentive commenting behavior (Definition~\ref{def:rationalinattention}) exists in the majority of YouTube user groups.
\item Our decision test is robust- The segments that do not pass the rational inattention test (Theorem~\ref{thrm:decisiontestrationalinattention}) are close to satisfying the inequalities in (\ref{eqn:MILPagentutility}).
\item The utility function constructed from the rational inattention test (Corollary~\ref{corl:MILP}) can be used to accurately predict commenting behavior (with $83\%$ accuracy). 
\end{compactenum}
%To construct $\mathcal{D}$,
\begin{figure}
	\begin{minipage}[c]{0.35\textwidth}
		\includegraphics[width=0.97\textwidth]{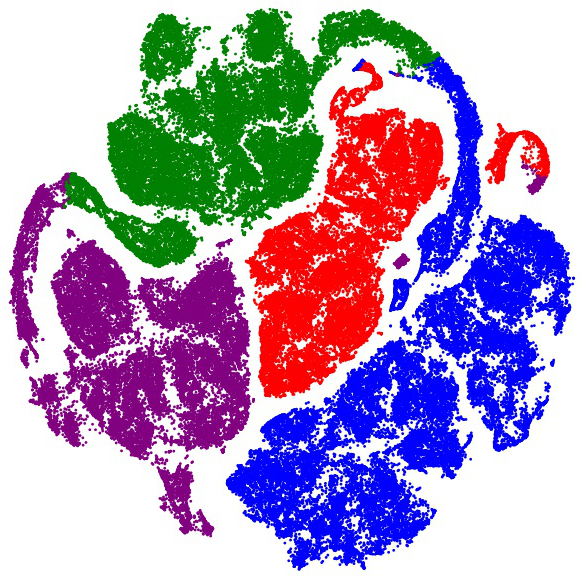}
	\end{minipage}\hfill
	\begin{minipage}[c]{0.65\textwidth}
		\caption{\small{t-SNE visualization~(\cite{maaten2008visualizing}) of the latent space representation of title and thumbnail of YouTube videos constructed using Algorithm~\ref{alg:framingassociation_1} for the YouTube videos contained in the dataset $\mathcal{D}$. As seen, the t-SNE representation with $4$ frames (each indicated by a different color) indicates that the videos are sufficiently separated in the latent space.}}	\label{fig:tsneunqiueframes}
	\end{minipage}
\end{figure}
The results presented in Sec.\ \ref{sec:DeepClustering} and Sec.\ \ref{sec:CatDissect} of this paper can be reproduced using the code and datasets that we have uploaded to a public GitHub repository:~\url{https://github.com/KunalP117/YouTube\_project}.
\subsection{YouTube Dataset and Model Parameters}
\label{sec:YTParameters}
The massive YouTube dataset that we analyze comprises $140,000$ videos across 25,000 channels from April 2007 to May 2015. We index the videos by $t=\{1,2,\dots T\}$, where $T=140,000$. The dataset contains the view counts, comment counts, likes, dislikes, thumbnail, title, and category of each video. To relate to our main Theorem~\ref{thrm:decisiontestrationalinattention}, we define the following:\newline
\textit{$1$. Agent}: Group of users (See Appendix~\ref{apdx:youtube} for more details on the number of users per group) interacting with videos in each video segment. Each agent is associated with a decision problem (either a frame or category as defined below). \newline 
\textit{$2$. State ($x_t$)}: In the YouTube dataset, the state $x_t$ of each video is the viewcount $1$ day after the video was published\footnote{Alternatively, we could have used the number of subscribers when the video was published as the state. Our extensive data analysis in \cite{HAK17} shows that the viewcount after 1 day and the initial subscriber count are the two highest sensitivity features for estimating the viewcount 14 days after the video is published.}. Specifically, state $x_t=1$ is high viewcount (more than $10,000$ views) and $x_t=2$ otherwise.\newline
\textit{$3.$ Action ($a_t$)}: In the YouTube dataset, the associated action $a_t$ is related to the overall commenting behavior\footnote{By overall commenting behavior in YouTube, we mean both the comment count and the video ratings (likes and dislikes). Another term used in the literature (\cite{KH17}) is ``user engagement''.} of the agents, which is computed using the comment counts, like count, and dislike count $2$ days after the video is published. 
The possible actions are: $a_t=1$ denotes low comment count with negative sentiment, $a_t=2$ denotes low comment count with neutral sentiment, $a_t=3$ denotes low comment count with positive sentiment, $a_t=4$ denotes high comment count with negative sentiment, $a_t=5$ denotes high comment count with neutral sentiment, and $a_t=6$ denotes high comment count with positive sentiment. Here negative sentiment occurs if the difference between the like count and dislike count is less than $-25$, neutral sentiment occurs if the difference lies between $-25, 25$, and positive sentiment occurs if the difference is greater than $25$. A low comment count is said to occur if there are less than $100$ comments, otherwise the comment count is defined to be high. If the user searches for a video, or directly goes to the video webpage,  then  the  thumbnail image and  viewcount are the first pieces of information that the user sees. The user needs to either click on the thumbnail or scroll down the specific video webpage (both actions affect the video's viewcount) in order to post their comments.
%\textcolor{blue}{Also, we assume that commenting behavior on the video webpage appearing after $2$ days depends on the existing comments, likes and dislikes.}
%In YouTube, videocount causes the commenting behavior since videos need to be viewed first before users can comment or rate the video. 
%\textcolor{blue}{To reduce this effect, we assume that the selected commenting behavior (action) is only dependent on the state (video viewcount after $1$ day) and decision problem (as discussed below).}
%Thus, we assume that the viewcount of the video after $1$ day is the independent variable which determines the commenting behavior $2$ days after the video is published (actions), in accordance with our definition of state and action. 
\newline
\textit{$4$. Frame ($f_t$)}: The frame $f_t$ of a video refers to the segment each video in the YouTube dataset is categorized into by Deep Embedded Clustering based on the video's framing information and is indexed as $1...N$. 
\begin{figure}[h]
	\centering
	\vspace{-9pt}
	\includegraphics[width=0.75\textwidth]{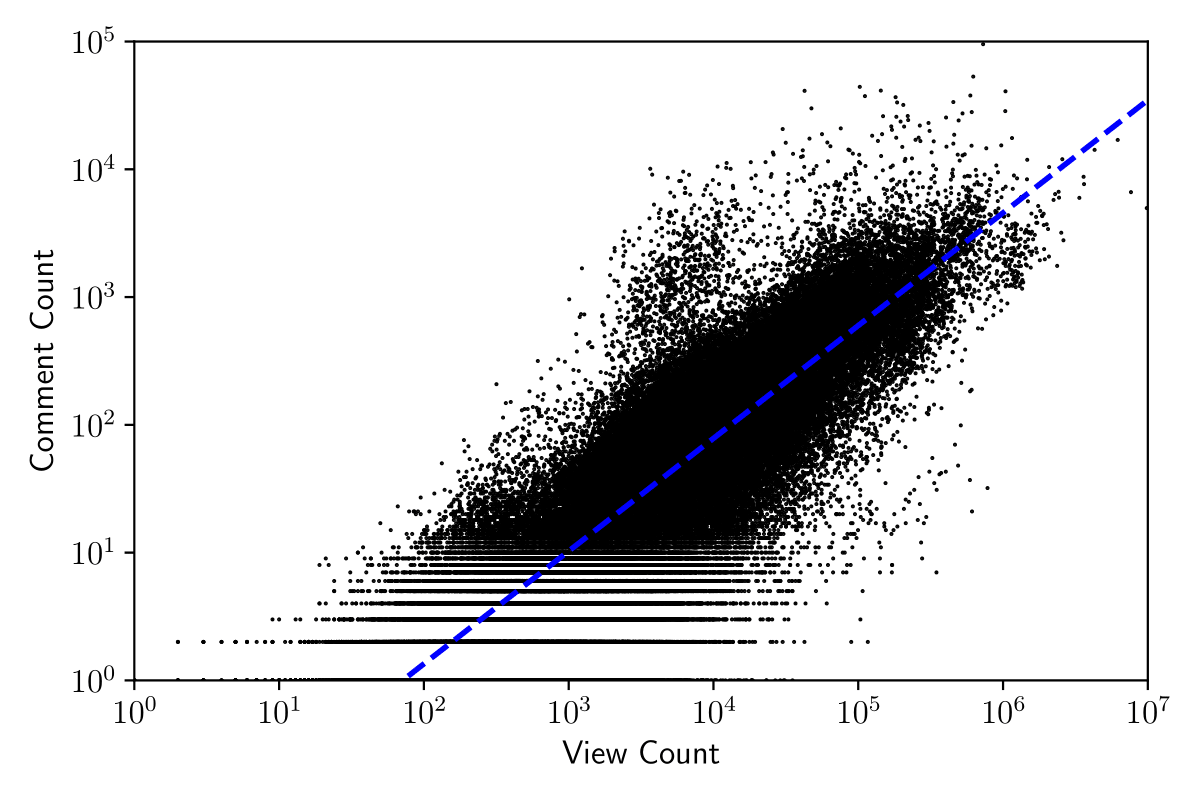}
	\vspace{-10pt}
	\caption{The comment count vs.\ viewcount follows a power law with power coefficient of $0.88$, namely, $c=0.023\times v^{0.88}$ in the YouTube dataset $\mathcal{D}$. So, the comment count increases sub-linearly and the rate of change of commenting decreases with increasing viewcount.}
	\label{fig:commentsvsviews}
\end{figure}The framing information for each video is comprised of the video's thumbnail and title. Specifically, we use a $40\times80$ pixel color image to represent the thumbnail (which is a resized version of the native $246\times138$ pixel thumbnails used in YouTube). For the title, we only include the first 8 words of the title in the framing instance $f_t$ (over 90\% of the videos have a title of length 8 words or less).\newline
\textit{$5$. Decision problem  ($k$)}: $(i)$ When data is partitioned by framing (via deep clustering methods of Sec.\ \ref{sec:FrameingInformationDeepLearning}), for each frame, a decision problem corresponds to one of two types the video belongs to - gaming videos or non-gaming videos (Sec.\ \ref{sec:DeepClustering}).
%the decision problem corresponds to one of two categories the video belongs to in each frame- gaming or non-gaming (Sec.~\ref{sec:DeepClustering}). 
$(ii)$ When data is partitioned by the video category (based on Sec.\ \ref{sec:CatDissect}), the decision problem is specified by the category each video belongs to. Each video contains a category index $k\in\{1,\dots,18\}$ representing the specific YouTube category descriptor of the video. Fig.~\ref{fig:categoryandviewcount} in Appendix~\ref{apdx:youtube} lists each video category with the total number of views. Note that the video categories ``Unavailable'' or ``Removed'' are videos flagged by YouTube as being suspected of violating YouTube's video policies\footnote{Refer to~\url{https://www.youtube.com/yt/about/policies/\#community-guidelines} for details}.\newline
\textit{$6.$ Information Acquisition Cost}: We consider the following parameterized costs of acquisition of attention function $\alpha_k$ used by the $k^{th}$ Bayesian agent~(\ref{eqn:attentionmaximization}): $(i)$ General cost~(\ref{eqn:ordinalcost}), $(ii)$ R{\'e}nyi mutual information cost~(\ref{eqn:renyimutualinformation}), $(iii)$ Shannon mutual information cost~(\ref{eqn:renyimutualinformation})

To give additional visual insight (although tangential to the paper), Fig.\ \ref{fig:commentsvsviews} displays the comment count vs.\ viewcount of $140,000$ videos comprising the dataset $\mathcal{D}$. As shown, a power law yields approximately $2\%$ smaller residual root mean squared error than a linear fit. It can be seen that while the associated comment count increases with view count, the rate of change of commenting decreases with increasing viewcount. Of course, this paper and the data analysis below focus on the different issue of how to determine a Bayesian utility function that rationalizes comments for groups of users given the viewcount.

\subsection{YouTube Data Analysis 1: Deep Clustering Approach}
\label{sec:DeepClustering}
\begin{figure}[h]
	\centering
	\vspace{-9pt}
	\includegraphics[width=1\textwidth]{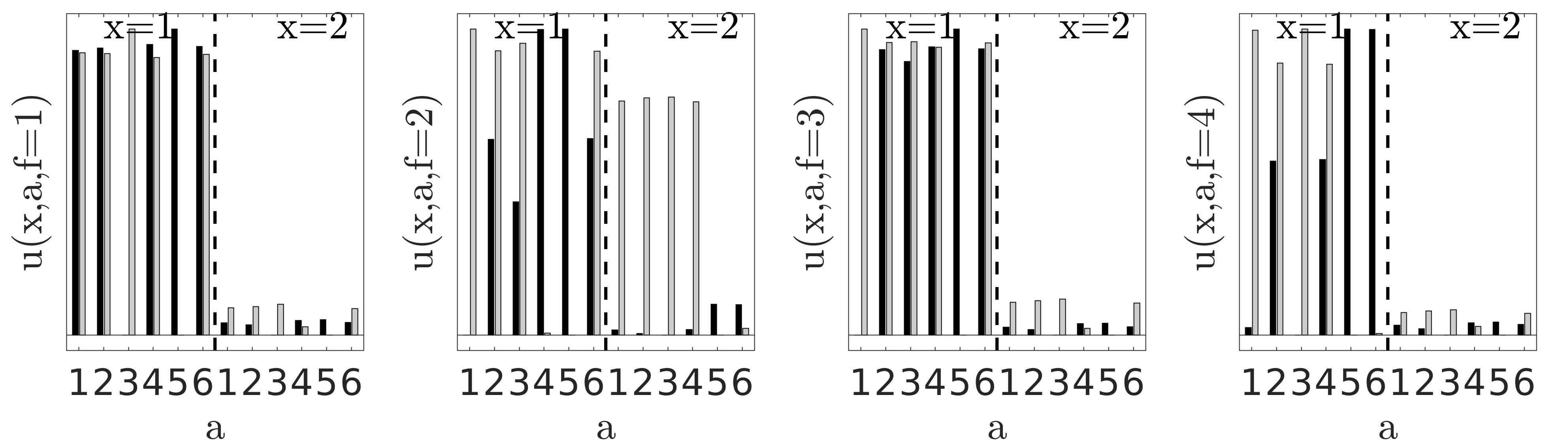}
	\vspace{-10pt}
	\caption{Utility functions $\{u_k(x,a,f)\}_{k=1}$ of the rationally inattentive agents with a general cost for information acquisition for each of the four unique frames are plotted in Fig.~\ref{fig:YouTubeutilitygeneralcost}. The utility is constructed by evaluating the mixed-integer linear feasibility problem (\ref{eqn:MILPagentutility}) for the YouTube dataset $\mathcal{D}$. $x$ represents the state, $a$ the possible actions, $f$ the frame, and the decision-problem $k$ indicates the most popular category (black bars) and the other categories (gray bars). 
		%That utility functions illustrating commenting preferences can be constructed for YouTube videos and that 
		The main conclusions are that a utility function can be constructed that explains the commenting preferences of users, which also indicates that a typical YouTube commenter has a higher preference (utility) to comment on videos with a higher viewcount (popularity metric) for all actions $a\in\{1,2,3,4,5,6\}$. Recall that these actions reflect the comment count and sentiment of comments.}
	\label{fig:YouTubeutilitygeneralcost}
\end{figure}
Our first approach for analyzing the YouTube dataset is user centric: it involves constructing frames over which preference ordering for nature of commenting behavior stays invariant using deep embedded clustering via Algorithm~\ref{alg:framingassociation_1}. Recall that the frame $f_t$ of a video refers to the segment each video in the YouTube dataset is categorized into by Deep Embedded Clustering based on the video's extrinsic framing information.
Algorithm~\ref{alg:framingassociation_1}  maps the high dimensional title and thumbnail space to one of $N$ unique frames. Fig.\ref{fig:tsneunqiueframes} depicts a two stage dimension reduction described as follows: The raw high dimensional vectors comprising video description and thumbnail are first mapped to a latent embedding space (of dimension 200) as part of Algorithm~\ref{alg:framingassociation_1}. 
%Using a similar Euclidean proximity preserving algorithm used in the auto-encoder (\cite{}),
The points in the latent embedding space are further projected to a 2-D space for better visualization (t-distributed Stochastic Neighbor Embedding~(\cite{maaten2008visualizing})). Each of the four colors in the figure represents a distinct frame.
% using the t-Distributed Stochastic Neighbor visualization.
Choosing $N=4$ ensures  each video is sufficiently isolated to a particular frame; less than
$3\%$ of videos are classified ambiguously  in terms of frames.
The most popular category of videos in the YouTube dataset is ``Gaming'' which comprises $44\%$ of all the videos. Two decision-problems are considered within each frame of the dataset. The first is $k=1$ which is associated with all videos that have category ``Gaming'', while decision problem $k=2$ results for videos that are not associated with the ``Gaming'' category. 
%In each group of videos pertaining to a single frame, we further partition the videos into two- one each for the gaming and non-gaming category.

\textbf{\textit{Data Analysis Results}}:~We apply the rational inattention test (\ref{eqn:MILPagentutility}) from Corollary~\ref{corl:MILP} to each of the preference invariant frames in Fig.~\ref{fig:tsneunqiueframes}. Our main conclusion is that the commenting behavior of users in  YouTube is consistent with rational inattention for a general cost constraint~(\ref{eqn:ordinalcost}). The ordinal utility of the users in each unique frame (Fig.~\ref{fig:tsneunqiueframes}) is provided in Fig.~\ref{fig:YouTubeutilitygeneralcost}. It can be inferred from the constructed utility function in Fig.~\ref{fig:YouTubeutilitygeneralcost} that users prefer to comment on videos that are expected to have a higher popularity compared to videos with lower popularity.

The associated utility however provides no clear preference ordering between the popularity of the video and the associated commenting behavior. This suggests that YouTube user groups in each frame are rationally inattentive with respect to a general cost. If we impose the R{\'{e}}nyi Mutual Information cost constraint~(\ref{eqn:renyiform}), we find that only the commenting behavior in frame $f=4$ is rationally inattentive. It was also found that frame $4$ 
satisfies rational inattention for a Shannon Mutual Information cost constraint~(\ref{eqn:renyiform}). Thus, the user groups in frame $4$ satisfy utility maximization for all information acquisition costs.

\subsection{YouTube Data Analysis 2 : Category Dissection Approach}
\label{sec:CatDissect}

We now describe a second approach for analyzing the YouTube dataset; this approach is data centric and uses YouTube video categories instead of frames. The diversity of videos in YouTube is immense, and it is natural to exploit this diversity for prediction of attributes in YouTube commenting behavior. Hence, we now analyze YouTube data by partitioning groups of videos based on their category. The aim is to determine if a utility function rationalizes each category of YouTube videos. 
%Through this approach, we can comment about rationality regarding commenting behavior in specific YouTube user groups, thereby greatly increasing the granularity of prediction.
Categories in YouTube (eg. News, Gaming, Music etc.) are numbered from $1 - 18$ (See Fig.~\ref{fig:categoryandviewcount} for the full listing). 
%This approach is in contrast to the deep clustering method in Sec.~\ref{sec:DeepClustering} where the YouTube data was partitioned based on the framing information of videos. 
The granularity of the analysis and prediction in this section is much finer than the analysis in Sec.\ \ref{sec:DeepClustering}. As discussed in Appendix~\ref{apdx:youtube}, the video categories have mean numbers of users ranging from $149$ to $4596$  for high viewcount (greater than $10000$) videos and $8$ to $1801$ for low viewcount videos (less than $10000$).

In the current formulation, we apply the rational inattention test (Corollary~\ref{corl:MILP}) on each distinct pair of video categories $(i,j)$ from the existing category index set. This implies the number of decision problems $K=2$ in~(\ref{eqn:bayesianexpectedutility}). Each decision problem corresponds to the distinct video categories $i$ and $j$ and the variable $f$ in Theorem~\ref{thrm:decisiontestrationalinattention} indexes the video category pair $(i,j)$.
The main findings of this section are:
\begin{compactenum}
\item Rationally inattentive commenting behavior holds for approximately $56\%$ of the video categories in YouTube.
\item Our decision test for rational inattention is robust to parameterization in information acquisition cost- with a small perturbation (quantified precisely below), every user group can be rationalized by a utility function irrespective of the information acquisition cost.
\item We construct utility function for specific video categories in YouTube and use these utility functions to predict (with $83\%$ accuracy) commenting behavior in those categories.
\end{compactenum}
\subsubsection{YouTube Dataset Analysis for Utility Maximization with General Cost Constraint}
\label{sec:prim_test}
%We fix the number of decision problems to test NIAC and NIAS on to be equal to $2$ and consider every pairwise combination of all possible categories (a total of $18 \choose 2$ = $153$ pairs).
%We test categories in YouTube for utility maximization with rational inattention under the following cost constraints: \\
%(i) General Cost (\ref{eqn:ordinalcost}) (ii) R\'{e}nyi Mutual Information Cost (\ref{eqn:renyimutualinformation}) (iii) Shannon Mutual Information Cost (\ref{eqn:renyimutualinformation}).
%\textbf{\textit{Data Analysis Results:}} 
Utility maximization for R\'{e}nyi/Shannon Mutual Information cost implies utility maximization with the unconstrained general cost. The results are displayed in Table~\ref{tab:PrimaryTesting} where it is shown that approximately $56\%$ of categories satisfy the general cost. $90\%$ of videos in the YouTube dataset belong to this set of categories.
For R{\'e}nyi Mutual Information cost~(\ref{eqn:renyiform}), we see that the number of pairs of video categories satisfying utility maximization increases with $\beta$.

 %\textcolor{red}{NEEDS CHANGES!!!!! - discuss implication, draw reference from Ng, talk about lesser structure, and rational model, and make changes in the letter too}
\subsubsection{Robustness Of Rationally Inattentive Utility Maximization Test}
\label{sec:RobustnessCategoryDissection}
 A natural robustness question is: For the categories of YouTube data that failed the utility maximization test of Theorem 1, how close are these categories to passing the utility maximization test?
 Similarly, for the categories in YouTube passing the utility maximization test, how far are these categories from failing the test?
 The purpose of this section is to define three robustness measures: one for the video categories in YouTube data that do not satisfy Theorem 1 and two for the categories in YouTube which do satisfy Theorem~\ref{thrm:decisiontestrationalinattention}. The main result below is that for the general cost for information acquisition~(\ref{eqn:ordinalcost}), these robustness measures are large for video category pairs that satisfy the utility maximization test and are relatively small for the category pairs that fail the test, for all $18$ video categories in YouTube. This implies that the commenting behavior in video categories that satisfy utility maximization do so by a large margin; and the video categories that fail utility maximization do so by only a small margin. Thus one can conclude that the commenting behavior across \textit{all} video categories in YouTube is approximately rational.
 \begin{table} 
 	%\begin{adjustbox}{width=0.7\columnwidth,center}
 	\begin{center}
 		\small{
 			\scalebox{0.8}{
 				\begin{tabular}{|m{4.45cm}|m{4.8cm}|m{4.7cm}|}
 					\hline
 					Information Acquisition Cost&Number of category pairs from $18\choose 2$ pairs that satisfy rational inattention decision test~(Corollary~\ref{corl:MILP}) & Set of categories\\
 					\hline
 					General~(\ref{eqn:ordinalcost})&\center{45}& $\{1,5,7,8,9,10,11,12,14,17\}$ \\
 					\hline
 					R\'{e}nyi Mutual Information~(\ref{eqn:renyiform})& &\\	
 					$\beta \in (0,0.7)$ &\center{$0$}& $\{\}$\\
 					$\beta \in [0.7,0.83)$ &\center{$2$}& $\{11,12\}$\\
 					$\beta \in [0.83,1)$ &\center{$3$}& $\{11,12,17\}$\\
 					\hline
 					Shannon Mutual Information~(\ref{eqn:renyiform})&\center{$2$}&$\{11,17\}$\\
 					\hline
 				\end{tabular}
 		}}
 		\caption{ Categories of YouTube videos where the commenting behavior satisfies Bayesian utility maximization behavior }
 		\label{tab:PrimaryTesting}
 	\end{center}
 \end{table}
% all video categories in YouTube is approximately rational.
 %We show that for all $18$ YouTube categories, these robustness measures are small implying that the commenting behavior across all user groups in YouTube is approximately consistent with utility maximization behavior.
%For each of the information cost structures in consideration, it is desirable to know whether the data that does not satisfy the rational inattention test in Theorem~\ref{thrm:decisiontestrationalinattention} is close to satisfying it.
%showing utility maximization behavior and if so, can we quantify it? 
%If yes, then we can comment in a richer sense about YouTube data rather than just a yes or no regarding utility maximization.
%If so, then we can assert that the commenting behavior is approximately consistent with utility maximization behavior. 
%There might be certain categories which violate (\ref{eqn:MILPagentutility}) to a "lesser" extent than others and consequently, closer to satisfying utility maximization. Therefore, it is relevant to evaluate the robustness of YouTube data to the rational inattention decision test (\ref{eqn:MILPagentutility}). 
%In the context of Theorem~\ref{thrm:decisiontestrationalinattention}, the equivalent question is:  how close is the data to satisfying the mixed-integer linear constraints (\ref{eqn:MILPagentutility}) and the KKT conditions in Theorem~\ref{thrm:reynicost}. Thus, we wish to determine the minimum perturbation needed to satisfy utility maximization behavior.
Recall from Sec.\ \ref{sec:YTParameters} that in our YouTube data analysis, the state space $\mathcal{X}=\{1,2\}$ and action space $\mathcal{A}=\{1,2,3,4,5,6\}$. Let $C=\{(i,j)|i\neq j,i,j\in \{1,2..,18\}\}$ denote the set of all possible distinct pairs of YouTube video categories. 
Based on Table~\ref{tab:PrimaryTesting}, denote the set of video category pairs that satisfy utility maximization for the general cost~(\ref{eqn:ordinalcost}) as:
\begin{equation*}
 C_{GC}=\{(i,j)|i\neq j,i,j\in\{1,5,7,8,9,10,11,12,14,17\}\}
\end{equation*}
%first we divide the set of distinct pairs of YouTube categories as foll: 
%let $C_{GC}=\{(i,j)|i\neq j,i,j\in\{1,5,7,8,9,10,11,12,14,17\}\}$ denote the set of video category pairs that satisfy utility maximization for the general cost structure and
%Let $C=\{(i,j)|i\neq j,i,j\in \{1,2..,18\}\}$ denote the set of all possible distinct pairs of YouTube video categories. 
%Also define $U_k (X,A)$ to be the utility vector of the $k^th$ decision problem with the $(x\times|A|+a)-$th element of $U_k (X,A) = u_k(x,a),x\in \{1,2\}, a \in \{1,2,3,4,5,6\}$.
%We show that for all $18$ YouTube categories, these robustness measures are small implying that the commenting behavior across all user groups in YouTube is approximately consistent with utility maximization behavior.
%We test the data on two key aspects in the following order: $(i)$ \textit{Bayesian Expected Utility Maximization} $(ii)$ \textit{Fitting a Functional form to Information Cost Structure}. 
We now introduce the three robustness metrics, two for $C_{GC}$ and one for $C\backslash C_{GC}$ as follows:\newline
{\bf 1. Robustness Metric ($\mathcal{R}_1$) for $C \backslash C_{GC}$}: For pairs of video categories that do not satisfy utility maximization, the robustness measure $\mathcal{R}_1$ computes how close these categories are to satisfying utility maximization. Define $\mathcal{R}_1$ for $C \backslash C_{GC}$ as 
%define the robustness metric $\mathcal{R}_1$ to be:
\begin{align} 
&\mathcal{R}_1 (i,j) = \min \frac{|\epsilon|}{\sqrt{(\sum_{x\in X} \sum_{a\in A} u_i(x,a)^2+u_j(x,a)^2)/2}}  \quad \forall (i,j)\in C\backslash C_{GC}\nonumber\\
&\sum_{x\in\mathcal{X}}p_{k}(x|a)[u_{k}(x,b)-u_{k}(x,a)] \leq \epsilon \quad \forall a,b\in\mathcal{A}, \forall k \in \{i,j\} \quad\label{eqn:epsilon}\\
&(G_{i,j} + G_{j,i})-(G_{i,i}+G_{j,j})\leq \epsilon \nonumber, 
\end{align}
where $G_{k,w}$ is as defined in (\ref{eqn:attentionmaximizationthrm}).
Similar to the perturbation metric used in~\cite{VAR85}, note that $\mathcal{R}_1$ is the minimum relaxation for a pair of categories $(i,j)$ to satisfy NIAC and NIAS in~(\ref{eqn:MILPagentutility}). Note that $\mathcal{R}_1$ is scale invariant; it is normalized by the average Euclidean norm of the utility vectors for different decision problems.

{\bf 2. Robustness metric ($\mathcal{R}_2$) for $C_{GC}$}: For pairs of video categories that satisfy utility maximization, what is the minimum perturbation so that they fail the utility maximization test? Define the robustness metric $\mathcal{R}_2$ for $C_{GC}$ as: 
\begin{align} 
&\mathcal{R}_2 (i,j) = \max \frac{|\epsilon|}{\sqrt{(\sum_{x\in X} \sum_{a\in A} u_i(x,a)^2+u_j(x,a)^2)/2}}  \quad \forall (i,j)\in  C_{GC}\nonumber\\
&\sum_{x\in\mathcal{X}}p_{k}(x|a)[u_{k}(x,b)-u_{k}(x,a)] + \epsilon \leq 0 \quad \forall a,b\in\mathcal{A}, \forall k \in \{i,j\} \quad \nonumber\\
&(G_{i,j} + G_{j,i})-(G_{i,i}+G_{j,j})+ \epsilon \leq 0 \nonumber, 
\end{align}
where $G_{k,w}$ is as defined in (\ref{eqn:attentionmaximizationthrm}). Note that in contrast to $\mathcal{R}_1$ defined above, $\mathcal{R}_2$ denotes the maximum perturbation/margin for a pair of categories $(i,j)$ such that NIAC and NIAS in~(\ref{eqn:MILPagentutility}) still hold. Also $\mathcal{R}_2$ is normalized by the average Euclidean norm of the utility vectors for different decision problems. Alternatively, $\mathcal{R}_2$ can be understood as the minimum perturbation needed for the YouTube category pairs $(i,j) \in C_{GC}$ to fail the utility maximization test for the general cost in Corollary~\ref{corl:MILP}. 

{\bf 3. Robustness Metric ($\mathcal{R}_3$) for $C_{GC}$}:
Finally, amongst the video category pairs that satisfy general cost utility maximization, how close are they to satisfying the more structured Shannon/R{\'e}nyi utility maximization? Define the robustness metric $\mathcal{R}_3$ for $C_{GC}$ as:
%For each $(i,j)\in C_{GC}$, we aim to find the minimum perturbation needed to satisfy the Shannon and R{\'e}nyi mutual information cost constraints. It is equivalent to satisfying~(\ref{eqn:KKT_1}),(\ref{eqn:KKT_2}) which can be written in vector form as:
%\mathcal{R}_2$ for a pair $(i,j)\in C_{GC}$ is defined as follows:
\begin{align}
& \mathcal{R}_3 (i,j) = \frac{1}{|K|}
\min \sum_{k= 1}^K \frac{\norm{\epsilon_{k}}^{2}}{\norm{\nabla I^{\beta}(\mu,\alpha_{k}}^2} \quad \forall (i,j)\in C_{GC}\label{eqn:R_2}\\
& [u_k(x,a), x\in \mathcal{X},a\in \mathcal{A}]^{T} = \lambda_{1,k} (\nabla I^{\beta}(\mu,\alpha_{k})-\epsilon_{k})-\lambda_{2,k}[1 1 1 ..1]^{T} \quad \forall k\in \{i,j\},\label{eqn:R2_KKT}\\
&\sum_{x\in\mathcal{X}}p_{k}(x|a)[u_{k}(x,b)-u_{k}(x,a,f)] \leq 0 \quad \forall a,b\in\mathcal{A} \quad \forall k\in \{i,j\}, \nonumber\\
&(G_{i,j} + G_{j,i})-(G_{i,i}+G_{j,j})\leq 0, \nonumber\\
& \lambda_{1,k} > 0, K=\{i,j\}\nonumber,
\end{align}
where $\epsilon_k$ is a gradient perturbation vector for category $k$, $G_{k,w}$ is as defined in (\ref{eqn:attentionmaximizationthrm}), and~(\ref{eqn:R2_KKT}) is the perturbed version of~(\ref{eqn:KKT_1}),~(\ref{eqn:KKT_2}), which in vector form can be written as:
\begin{equation}
[u_k(x,a), x\in \mathcal{X},a\in \mathcal{A}]^{T} = \lambda_{1,k} (\nabla I^{\beta}(\mu,\alpha_{k}))-\lambda_{2,k}[1 1 1 ..1]^{T}, \label{eqn:vec_KKT}
\end{equation}
where $k$ indexes the $k^{th}$ decision problem. 
We aim to find the minimum perturbation needed for each category pair in $C_{GC}$ to be utility maximizers with constraints on the information acquisition cost (Shannon and R{\'e}nyi Mutual Information~(\ref{eqn:renyiform})). 
For computing $\mathcal{R}_3$, the Euclidean norm of the minimum perturbation vector $\epsilon_k$ for each decision problem $k$ is normalized by its gradient vector and averaged over all decision problems.
\newline \newline
\newline
\textbf{\textit{YouTube Dataset Robustness Analysis Results}}: With the above three robustness measures, we now analyze the YouTube dataset.
\newline
\textit{(i) General cost~(\ref{eqn:ordinalcost})}:
The average $\mathcal{R}_1$ over all category pairs $(i,j)\in C\backslash C_{GC}$ was found to be $1.2\times 10^{-3}$. Moreover, $\mathcal{R}_1\leq 9\times 10^{-4}$ for $68\%$ of category pairs $(i,j)\in C\backslash C_{GC}$. These results show that a small perturbation in data for categories in the set $\{2,3,4,6,13,15,16,18\}$ ensures that they satisfy utility maximization for the general cost constraint. 

\begin{figure}
	%[h]
		\centering
		%\vspace{-1pt}
		\includegraphics[width=0.75\textwidth]{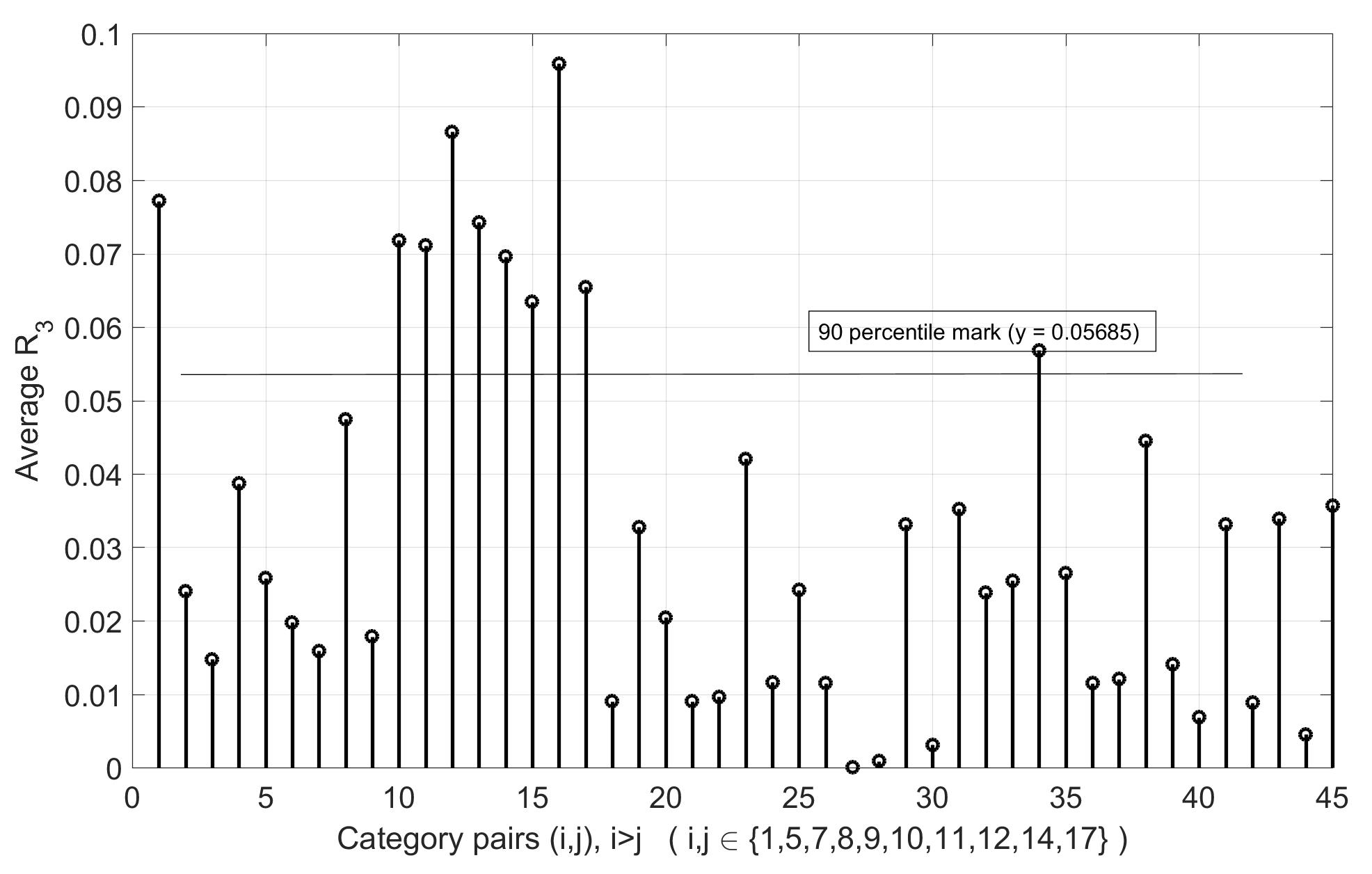}
		%\vspace{-10pt}
		\caption{The vertical axis depicts $\mathcal{R}_3 (i,j)$~(\ref{eqn:R_2}) for each $(i,j)\in C_{GC}$. With the $90$-th percentile of robustness measure $\mathcal{R}_3$ defined in~(\ref{eqn:R_2}) approximately equal to $0.06$, we conclude that all the user groups for categories in the set $\{1,5,7,8,9,10,11,12,14,17\}$ are approximately consistent with utility maximization for Shannon mutual information cost. This set comprises 90\% of videos in the YouTube dataset.}
		\label{fig:RobustnessShannon} 
\end{figure}

\begin{figure}
	%[h]
	\centering
	%\vspace{-9pt}
	\includegraphics[width=0.75\textwidth]{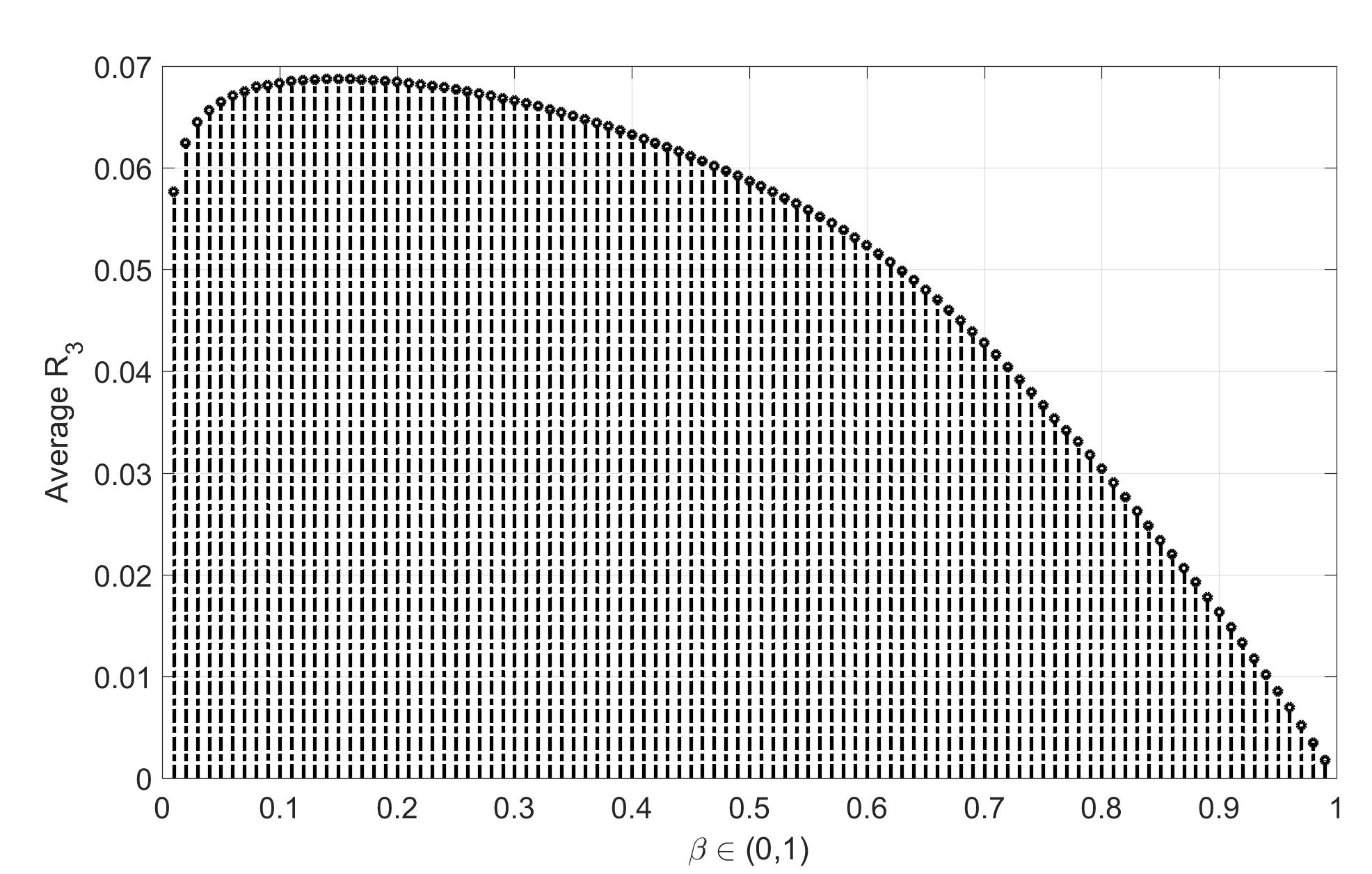}
	%\vspace{-10pt}
	\caption{Robustness measure $\mathcal{R}_3$~(\ref{eqn:R_2}) averaged over $C_{GC}$ for parameter $\beta \in (0,1)$ in R{\'e}nyi mutual information is plotted on the vertical axis. 
	%The perturbations for each value of $\beta$ are averaged over $\mathcal{K} (=2)$ decision problems and the pairs of categories $(i,j)~i>j$ that satisfy utility maximization without cost constraints.  The average norm of normalized perturbation is small and decreases with increasing $\beta$, thus showing that the data needs to be perturbed minimally to be consistent with utility maximization with R\'{e}nyi  information cost structure.
	The average $\mathcal{R}_3$ decreases for $\beta \geq 0.15$ implying that for each category in $C_{GC}$, a smaller perturbation is needed to satisfy utility maximization model with R{\'e}nyi mutual information cost for higher $\beta$. }
	\label{fig:RobustnessRenyi}
\end{figure}

In contrast, the average $\mathcal{R}_2$ over all category pairs $(i,j)\in C_{GC}$ was found to be $7\times 10^{-3}$, approximately $5$ times the average value of $\mathcal{R}_1$. $\mathcal{R}_2 \geq 5 \times 10^{-3}$ for $65\%$ of pairs of YouTube categories $(i,j)\in C_{GC}$. This result shows that the minimum perturbation required for category pairs in $C_{GC}$ to fail the utility maximization test is relatively large compared to the minimum perturbation needed for category pairs in $C \backslash C_{GC}$ to pass the test; this highlights that our utility maximization model for the general cost constraint is robust with respect to modeling commenting behavior in YouTube categories.
\newline \newline
\textit{(ii) Shannon mutual information cost~(\ref{eqn:renyiform})}:
$\mathcal{R}_3$ for $\beta=1$ over all category pairs $(i,j)\in C_{GC}$ are displayed in Fig.~\ref{fig:RobustnessShannon}. The highest value is $0.096$, with $90\%$ of the pairs $(i,j)\in C_{GC}$ having $\mathcal{R}_3 \leq 0.057$. Fig.~\ref{fig:RobustnessShannon} indicates that all categories in the set $\{1,5,7,8,9,10,11,12,14,17\}$ ($56\%$ of categories) are approximately consistent with utility maximization behavior for the Shannon mutual information cost.\newline
\textit{(iii) R\'{e}nyi mutual information cost~(\ref{eqn:renyiform})}: Fig.~\ref{fig:RobustnessRenyi} displays $\mathcal{R}_3$ averaged over $C_{GC}$, that is, \newline $\left(\frac{\sum_{(i,j)\in C_{GC}}\mathcal{R}_3 (i,j)}{|C_{GC}|}\right)$ for $\beta\in (0,1)$. The maximum averaged $\mathcal{R}_3$ recorded over $\beta\in (0,1)$ was found to be $0.069$ which indicates that for any $\beta\in (0,1)$, a small perturbation is sufficient to make user groups for categories in the set $\{1,5,7,8,9,10,11,12,14,17\}$ satisfy utility maximization for R{\'e}nyi mutual information cost. Interestingly, one can see that the average $\mathcal{R}_3$ decreases with $\beta$, implying that the R{\'e}nyi mutual information cost constraint fits the data better for a higher $\beta$, where the average value of $\mathcal{R}_3 \leq 0.01$ for $\beta \geq 0.9$.

%Both Shannon and R\'{e}nyi Mutual Information cost structures exhibit minimal deviation from optimal behavior. The deviation is negligible in either case which points to an intriguing aspect of YouTube data - is the YouTube data resilient enough to fit multiple cost structures well? 
According to our robustness analysis, the commenting behavior of user groups in all $18$ YouTube video categories approximately satisfies utility maximization for the general cost (\ref{eqn:ordinalcost}); for $56\%$ of video categories in YouTube, the user groups approximately satisfy rationally inattentive commenting behavior for all information acquisition costs, namely, general cost~(\ref{eqn:ordinalcost}) and R{\'e}nyi/Shannon mutual information cost (\ref{eqn:renyiform}) for all $\beta\in (0,1]$.
%one can expect a typical user to approximately exhibit rational inattention while commenting on a video, over a wide range of information acquisition costs. 
The next natural question is: can we predict what type of commenting behavior to expect given the viewcount of a video in a particular category? We explore this aspect in the following sub-section.

\subsubsection{Behavior Prediction Using Utility Function}
\label{sec:BehaviorPrediction}
The results in Sec.\ \ref{sec:DeepClustering}, \ref{sec:prim_test} and \ref{sec:RobustnessCategoryDissection} indicate that YouTube users fit the rationally inattentive Bayesian model remarkably well. The next step is to demonstrate how the rational inattention test in Corollary~\ref{corl:MILP} can be used to predict commenting behavior in specific video categories based on the videos' viewcount (high or low). The main result below is that the
comment count (high or low) in the YouTube dataset can be predicted correctly with $83\%$ accuracy.

We divided the YouTube dataset $\mathcal{D}$ into two parts - training data ($80\%$) and testing data ($20\%$). Using Corollary~\ref{corl:MILP}, utility estimates for different commenting behavior was constructed for users in each category of YouTube videos in the training data. Out of all categories ($1-18$), the set $\{1,5,8,9,10,11,12,14\}$ was found to be consistent with utility maximization for the general cost~(\ref{eqn:ordinalcost}); we can construct utility estimates using Corollary~\ref{corl:MILP} for categories belonging to this set. For the remaining categories, it was found from the robustness test in Sec.\ \ref{sec:RobustnessCategoryDissection} that the average minimum relaxation needed for utility maximization behavior is $9.4\times 10^{-4}$.

To quantify the user commenting behavior on the testing data, the \textit{Maximum a posteriori} (MAP) estimate, namely  ($\underset{a\in\mathcal{A}}{\arg\max}~p_{k}(x|a)$), is computed for each state $x\in \{1,2\}$ in every category $k$ which satisfied the decision test for utility maximization. Note that the MAP estimate is the maximum likelihood estimate of the action unless the prior over actions is uniform. Similarly on the training data, define the \textit{Maximum Utility Estimate} as $\underset{a\in\mathcal{A}}{\arg\max}~u_{k}(x,a)$, for each state $x\in\{1,2\}$ in every category $k\in\{1,5,8,9,10,11,12,14\}$.  
%The description of the state and action follows from Sec.~\ref{sec:RationalInattentionandUtilityMaximizationintheYouTubeSocialNetwork}

%The output of our test (Theorem~\ref{thrm:decisiontestrationalinattention}) yields set-valued estimates of the utility ($u_{k}(x,a)$) as a function of state and action in each of the $8$ categories~($k$), from which $\underset{a\in\mathcal{A}}{\arg\max}~u_{k}(x,a)$ (Maximum Utility Estimate) is computed for each category. 

Recall from Sec.\ \ref{sec:YTParameters} that our actions are numbered from $1-6$, with $1-3$ for low comment count, and the rest for high comment count. 
%Actions $1-3$ and $4-6$ differ in the  nature of the sentiment - positive, neutral and negative respectively. Hence, 
We define the error in commenting behavior (action) between prediction  (from training data) and observation (from testing data) in each state and category to be the 
%absolute difference 
Hamming distance between the MAP estimate (from testing data) and the maximum utility estimate (from the training data) -  $|\underset{a\in\mathcal{A}}{\arg\max}~u_{k}(x,a) -~\underset{a\in\mathcal{A}}{\arg\max}~p_{k}(x|a)) |$ as shown in Table.~\ref{tab:Prediction}~\footnote{If the rational inattention cost constraint~(\ref{eqn:attentionmaximization}) is omitted from Theorem~\ref{thrm:decisiontestrationalinattention}, then {\em all $18$ video categories in the YouTube dataset} pass the utility maximization test. However, the predictive capability of this less structured Bayesian utility model is substantially less accurate- the mean squared prediction error (in terms of Hamming distance) is $36\%$ higher ($3.85$) compared to the original model ($2.45$) as discussed in Table~\ref{tab:Prediction}. We thank an anonymous reviewer for suggesting this comparison.}.

Table~\ref{tab:Prediction} also shows that the nature of comment count (high or low) is identical for a particular state $x\in\{1,2\}$ and category $k$ in training and testing data if both MAP estimate and Maximum utility estimate belong to either $\{1,2,3\}$ or $\{4,5,6\}$.
%The estimated and actual actions for a particular state $x$ and category $k$ have the same comment sentiment if they are congruent modulo $3$.
From the results in Table~\ref{tab:Prediction}, atleast one of the two aspects of commenting behavior (nature of comment count (high or low) and sentiment) can be predicted correctly with $83.33\%$ accuracy in $16(8\times2)$ out of $36(18\times 2)$ sub-categories in YouTube.
%From the results in Table~\ref{tab:Prediction}, the nature of comment count (high or low) in the YouTube dataset can be predicted correctly with $83\%$ accuracy.

\begin{table}
\begin{center}
	\small
	\scalebox{1.00}{
		\begin{tabular}{|m{1.3cm}|m{1.3cm}|m{2.3cm}|m{1.3cm}|m{1.3cm}|m{2.3cm}|m{1.3cm}|}
	\hline
	Category&MAP estimate ($x=1$)& Max. Utility \newline estimate ($x=1$)&Error ($x=1$)&MAP estimate ($x=2$)&Max. Utility \newline estimate ($x=2$)& Error ($x=2$)\\
	\hline
	\hfil1&\hfil 5&\hfil5&\hfil0&\hfil3&\hfil1&\hfil2\\
	\hline
	\hfil5&\hfil5&\hfil5&\hfil0&\hfil5&\hfil2&\hfil3\\
	\hline
	\hfil8&\hfil5&\hfil5&\hfil0&\hfil1&\hfil3&\hfil2\\
	\hline
	\hfil9&\hfil6&\hfil5&\hfil1&\hfil3&\hfil2&\hfil1\\
	\hline
	\hfil10&\hfil5&\hfil5&\hfil0&\hfil3&\hfil2&\hfil1\\
	\hline
	\hfil11&\hfil5&\hfil5&\hfil0&\hfil6&\hfil3&\hfil3\\
	\hline
	\hfil12&\hfil5&\hfil4&\hfil1&\hfil3&\hfil2&\hfil1\\
	\hline
	\hfil14&\hfil6&\hfil5&\hfil1&\hfil3&\hfil1&\hfil2\\
	\hline
	\hfil17&\hfil5&\hfil2&\hfil3&\hfil3&\hfil1&\hfil2\\
	\hline
\end{tabular}}
	
%\caption{Validation Results : Observed vs Predicted}
%\label{tab:prediction}
\end{center}
\caption{Prediction Results for YouTube Commenting behavior: 
The mismatch between the commenting behavior in training and testing data is displayed for each state $x\in\{1,2\}$ (high or low viewcount) and each category which satisfies utility maximization. The mean squared prediction error in terms of Hamming distance error is $2.45$. In $83.33\%$ of these cases, the prediction error is at most $2$ units (in terms of Hamming distance) compared to maximum possible error of $5$ units. In $27.77\%$ of the cases the prediction error is zero, $61.11\%$ of the cases have an error $\leq 1$ unit and $83.33\%$ of the cases have and error $\leq 2$ units while the maximum error recorded is $3$ units. Thus the rationally inattentive utility maximization model provides an accurate predictor for YouTube commenting behavior.}
\label{tab:Prediction}
\end{table}
\section*{Discussion}
%Both methods of analysis show great similarity in the results and provide deep insights into the way comments results in Youtube videos. The approaches are primarily driven by the need to granularize the data - we wish to find estimates of the utility function in small subsets within the massive YouTube data so as to predict commenting behavior more precisely. 
 
The deep clustering approach in Sec.\ \ref{sec:DeepClustering} is user centric: it categorizes videos into frames based on the videos' extrinsic framing information. This approach is implemented via Algorithm~\ref{alg:framingassociation_1} which preserves "closeness"  between videos in both the high dimensional space (where the video thumbnail and description is decoded to) and the low dimensional space (achieved through deep embedded clustering) and achieves a granularity of $4$. In comparison, the category dissection approach of Sec.\ \ref{sec:CatDissect} is content centric; it exploits differing commenting behavior across YouTube categories~(\cite{SD10}) and separates them based on the videos' individual category, achieving a granularity of $18$.
% We could go on increasing the level of granularity, but are constrained by the limited amount of data at our disposal and the increasing deviation of the empirically calculated probabilities from the actual probabilities with granularity.

Based on extensive analysis of the YouTube dataset, our main conclusions are that users' commenting behavior (comment count and comment sentiment) is i) consistent with rational inattention, ii) depends on the framing information available and the category each video belongs to iii) users  prefer to comment on videos that are perceived to be popular. Since the action is also indicative of the perceived sentiment, one can predict how different videos will be perceived based on their popularity over different segments.
%\vspace{0.2cm}

 Note that our analysis was based on a one-to-one correspondence between the signals and actions which makes the action selection policy exactly the same as the attention function. 
%As can be seen in (\ref{eqn:appdx_attn_selcn}), 
Relaxing the one-to-one dependency would ensure a higher fraction of data to be consistent with rational inattention. In spite of having used a parsimonious representation of the attention function from the available data, we have unearthed promising insights into YouTube's video interaction dynamics.

 That deep embedded clustering adequately captures  framing information, that segregating videos by their categories enables the data to fit a rationally inattentive utility maximization model from which the preference based utility obtained with attention costs  rationalizes the YouTube dataset is remarkable. We were also able to predict accurately~($83\%$) the observed commenting behavior from the generated utility function.  
 %\vspace{0.3cm}
%Before deciding to view a particular video, an agent does not get to view the posted comments, thus making the decision to view a video (the likelihood of which is reflected by the state variable $x$ (viewcount) ) independent of the commenting behavior. Hence, any inferences drawn regarding the correlation between viewcount and commenting behavior may as well be extended to a causation conclusion, for eg. a high viewcount  "causes" a large number of comments. This is in contrast to social media like Facebook that allows the users to view a small fraction of the comments before they can play the video.
%From a behavioral economics point of view, the above results yield useful insight into user behavior in online social multimedia. There is also considerable scope to generalize the utility function estimation described in this paper to  stopping time problems involving partially observed Markov decision processes (\cite{Kri13,Kri16}).
%%%%%%%%%%%%%%%%%%%%%%%%%%%%%%%%%%%%%%%%%%%%%%%%%%%%%%%%%%%%%%%%%%%%%%%%%%%%%%%%%%

\section{Conclusion}
\label{sec:Conc} 
%\textcolor{red}{This paragraph is like scrambled eggs  - it needs a careful sequential structure as follows:  (i) The main ideas in this paper was ... such Bayesian revealed preferences addresses the deeper issue of ... compared to classical inverse RL  (ii) Understanding engagement is important because ..  (iii) The main result of this paper was Theorem xx which says that (iv) The key application if this paper was ...(v) Our main finding regarding YouTube was that ....}
This paper studied a novel class of inverse reinforcement learning methods for contextual bandit problems with behavioral economics constraints to learn and predict the commenting behavior of YouTube users. The main ideas in this paper involve Bayesian Revealed Preferences, Rational Inattention and Deep Embedded Clustering. Bayesian Revealed Preferences addresses the deeper issue of the existence of a utility function that rationalizes the given data compared to classical inverse reinforcement learning where the existence of such a function is implicitly assumed. Understanding commenting behavior (overall user engagement) in YouTube is important for modeling how humans interactively perceive social multimedia information, for YouTube partners like BBT and VEVO to increase their revenue, and for efficient content caching in wireless technologies like $5$G. 

The main result of this paper was Theorem~\ref{thrm:decisiontestrationalinattention} and Corollary~\ref{corl:MILP} which outline a decision test for utility maximization in Bayesian agents, and Theorem~\ref{thrm:reynicost} which provides conditions for satisfying additional behavioral economics based information acquisition constraints for the agent. The key application of this paper was to identify rationally inattentive YouTube user groups - groups that were consistent with utility maximization under specific information acquisition cost constraints - and show how the reconstructed utility with rational inattention constraints can be used to predict future commenting behavior in such groups accurately.
%In YouTube, videocount causes the commenting behavior since videos need to be viewed first before users can comment or rate the video. Thus, viewcount is the independent quantity which governs the commenting behavior.
Our main finding was that YouTube user groups are approximately rationally inattentive; users groups prefer to comment on videos that are popular (high viewcount); and the utility function constructed from the utility maximization decision test can be used to predict commenting behavior of these user groups. 
\vspace{0.5cm}

{\bf Reproducibility:} The computer programs and YouTube datasets needed to completely reproduce all the results in this paper can be accessed via the public GitHub repository:~ 
\small{\url{https://github.com/KunalP117/YouTube\_project}}.

 \newpage

 \appendix

 \section{Proofs}
 \label{appdx:proofs}
 \subsection{Theorem~\ref{thrm:decisiontestrationalinattention}} 
 	%The detailed proof is in \cite{CD15}. Here we present the key ideas.\newline
 {\it Proof of necessity of NIAS and NIAC:}	
 \begin{compactenum}
 	\item NIAS~(\ref{eqn:bayesianexpectedutility}): Consider any posterior belief $p(x|s)$ such that action $a\in \mathcal{A}$ is chosen by the agent i.e. $\sum_{x\in \mathcal{X}}p(x|s)(u(x,a)-u(x,b))\geq 0 $. 
 	The revealed posterior $p(x|a)$ (defined in~(\ref{eqn:bayesianexpectedutility})) is a stochastically garbled version of the actual posterior $p(x|s)$ i.e.,
 	\begin{equation*}
 	  p(x|a) = \sum_{s\in\mathcal{S}(\alpha)} \frac{p(x,a,s)}{p(a)}= \sum_{s\in\mathcal{S}(\alpha)}\frac{ p(a|s)p(s)p(x|s) }{p(a)} = \sum_{s\in\mathcal{S}(\alpha)} \frac{p(a,s)}{p(a)}p(x|s) = \sum_{s\in\mathcal{S}(\alpha)} p(s|a)p(x|s)
 	\end{equation*} 
 	where $\alpha$ is the attention function of the agent defined in~(\ref{eqn:utilitymaximization}). %The necessity proof of NIAS uses the fact that the revealed posterior is a garbled version of the actual posterior. 
 	Since the optimal action is $a$, the following holds from optimality for all $a\in\mathcal{A}$
 	\begin{equation*}
 	\begin{split}
 	& \sum_{x\in \mathcal{X}} p(x|s) (u(x,a)-u(x,b))\geq 0\\
 	\implies & \sum_{s\in \mathcal{S} } p(s|a) \sum_{x\in \mathcal{X}} p(x|s) (u(x,a)-u(x,b))\geq 0 \\
 	\implies & \sum_{x\in \mathcal{X}}\left( \sum_{s\in \mathcal{S}} p(s|a) p(x|s) \right)(u(x,a)-u(x,b))\geq 0 \\
 	\implies & \boxed{\sum_{x\in \mathcal{X}} p(x|a) (u(x,a) - u(x,b))\geq 0}
 	\end{split}
 	%\label{eqn:proof_NIAS}
 	\end{equation*}  
 	\item NIAC~(\ref{eqn:attentionmaximizationthrm}): Let $J(\alpha,\mathcal{A})$ denote the expected utility of the agent when using attention function $\alpha$ (defined in~(\ref{eqn:utilitymaximization})) for decision problem $\mathcal{A}$ (defined in Sec.\ \ref{sec:RationalInattentionandUtilityMaximizationintheYouTubeSocialNetwork}) i.e., 
 	\begin{equation*}
 	J(\alpha,\mathcal{A}) = \sum_{s\in \mathcal{S}(\alpha)} \left(\sum_{x\in \mathcal{X}} \mu(x)\alpha(s|x) \right)\left[ \max_{a\in \mathcal{A}} \sum_{x\in \mathcal{X}} p(x|s)u(x,a)\right]
 	\end{equation*}   
 	For any sequence of decision problems $\{\mathcal{A}_i\}_{i=1}^k$ 
 	%(recall the decision problems for the YouTube dataset $\mathcal{D}$ are defined in~Sec.\ \ref{sec:RationalInattentionandUtilityMaximizationintheYouTubeSocialNetwork}) 
 	and their optimal attention functions $\{\alpha_i\}_{i=1}^k$, we obtain the following relation from optimality using~(\ref{eqn:attentionmaximization})
 	\begin{equation}
 	\begin{split}
 	& J(\alpha_j,\mathcal{A}_j) - C(\mu,\alpha_j) \geq   J(\alpha_{j+1},\mathcal{A}_j) - C(\mu,\alpha_{j+1})\\ 
 	\implies & J(\alpha_j,\mathcal{A}_j) - J(\alpha_{j+1},\mathcal{A}_j) \geq C(\mu,\alpha_j) - C(\mu,\alpha_{j+1}) \\
 	\implies & \boxed{\sum_{j=1}^k J(\alpha_j,\mathcal{A}_j) - J(\alpha_{j+1},\mathcal{A}_j) \geq \sum_{j=1}^k C(\mu,\alpha_j) - C(\mu,\alpha_{j+1}) =0}\quad (\mathcal{A}_{k+1}=\mathcal{A}_1, \alpha_{k+1}=\alpha_1) 
 	\end{split}
 	\label{eqn:ineq_NIAC}
 	\end{equation}
 	The following equations establish the relationship between $J(\alpha_i,\mathcal{A}_j)$ and the observed expected utility $G_{i,j}$ defined in~(\ref{eqn:attentionmaximizationthrm})
 	\begin{equation}
 	\begin{split}
 	G_{i,j} = & \sum_{a\in\mathcal{A}_j} \sum_{x\in\mathcal{X}} p_i(x,a)u_j(x,a)\\
 	=& \sum_{a\in\mathcal{A}_j} \sum_{s\in\mathcal{S}(\alpha_i)}\sum_{x\in\mathcal{X}} p_i(x,a,s)u_j(x,a)\\
 	=& \sum_{a\in\mathcal{A}_j} \sum_{s\in\mathcal{S}(\alpha_i)} p_i(a|s)p_i(s) \sum_{x\in\mathcal{X}} p_i(x|s)u_j(x,a)\\
 	=& \sum_{s\in\mathcal{S}(\alpha_i)} p_i(s)\left[\sum_{a\in\mathcal{A}_j}p_i(a|s)\sum_{x\in\mathcal{X}} p_i(x|s)u_j(x,a)\right]\\
 	%=& \sum_{a\in\mathcal{A}_j} p_i(a) \sum_{x\in\mathcal{X}} \sum_{s\in\mathcal{S}(\alpha_i)} p_i(s|a)p_i(x|s)u_j(x,a)\\
 	%=& \sum_{a\in\mathcal{A}_j,s\in\mathcal{S}(\alpha_i)} p_i(a,s) \sum_{x\in\mathcal{X}} p_i(x|s)u_j(x,a)\\
 	%=&\sum_{s\in\mathcal{S}(\alpha_i)} p_i(s)  \left[\sum_{a\in\mathcal{A}_j} p_i(a|s)\sum_{x\in\mathcal{X}}p_i(x|s)u_j(x,a)\right]\\
 	\leq & \sum_{s\in\mathcal{S}(\alpha_i)} p_i(s)\left[ \max_{a\in\mathcal{A}_j}\sum_{x\in\mathcal{X}}p_i(x|s)u_j(x,a)\right] = J(\alpha_i,\mathcal{A}_j)\\
 	\implies & J(\alpha_i,\mathcal{A}_j) \geq G_{i,j}
 	\end{split}
 	\label{eqn:Blackwell}
 	\end{equation}
 	Clearly, equality holds above if $i=j$.
 	% 	Clearly, if $i=j$ above, then $J(\alpha_i,\mathcal{A}_i)=G_{i,i}$.
 Using~(\ref{eqn:Blackwell}), we further write~(\ref{eqn:ineq_NIAC}) as
 	\begin{equation*}
 	\begin{split}
 	%& J(\alpha_j, \mathcal{A}_j) = G_{j,j},\quad  J(\alpha_j, \mathcal{A}_l) \geq G_{j,l} \quad (j\neq l)\\
 	& \sum_{j=1}^k J(\alpha_j,\mathcal{A}_j) \geq J(\alpha_{j+1},\mathcal{A}_j) \implies \sum_{j=1}^k G_{j,j} \geq  \sum_{i=1}^k J(\alpha_{j+1},\mathcal{A}_j) \geq \sum_{i=1}^k G_{j,j+1}\\
 	\text{Hence, }&\boxed{ \sum_{j=1}^k G_{j,j} - G_{j,j+1} \geq 0} \quad (G_{k,k+1}=G_{k,1})
 	\end{split}
 	\end{equation*}
 \end{compactenum}
 {\it Proof of sufficiency of NIAS and NIAC:} \newline Given any decision problem, our assumption of a one-to-one mapping from the set of signals ($\mathcal{S}(\alpha)$) to the set of actions ($\mathcal{A}$) in Sec.\ \ref{sec:MutualInformationMeasuresandAttentionCost} implies $p(s|a) = 1$ for the signal corresponding to the action $a\in\mathcal{A}$ and $0$ elsewhere. Thus, we first show that the NIAS condition~(\ref{eqn:bayesianexpectedutility}) in Theorem~\ref{def:rationalinattention} implies~(\ref{eqn:utilitymaximization}) in Definition~\ref{def:rationalinattention}. 
 \begin{equation}
 \begin{split}
 \text{NIAS: }& \sum_{x\in\mathcal{X}}p(x|a)\left[ u(x,a)-u(x,b)  \right]\geq 0\quad \forall a,b \in \mathcal{A} \nonumber\\
 =& \sum_{s\in\mathcal{S}(\alpha)} \sum_{x\in\mathcal{X}}\frac{p(a,s,x)}{p(a)}\left[ u(x,a)-u(x,b)  \right]\geq 0\quad \forall a,b \in \mathcal{A}\nonumber\\
 =& \sum_{s\in\mathcal{S}(\alpha)} \frac{p(a|s)p(s)}{p(a)}\left[\sum_{x\in\mathcal{X}}p(x|s)\left[ u(x,a)-u(x,b)  \right]\right]\geq 0\quad \forall a,b \in \mathcal{A}\nonumber\\
 =&\sum_{s\in\mathcal{S}(\alpha)} p(s|a)\left[\sum_{x\in\mathcal{X}}p(x|s)\left[ u(x,a)-u(x,b)  \right]\right]\geq 0\quad \forall a,b \in \mathcal{A} \\
 =&\sum_{x\in\mathcal{X}}p(x|s_a)\left[ u(x,a)-u(x,b)  \right]\geq 0\quad \forall a,b \in \mathcal{A} \quad (p(s_a|a)=1)
 \end{split}
 \end{equation} 

 Define a square matrix $A_{K\times K}$ with elements $A_{i,j} = G_{i,j} $ (defined in~(\ref{eqn:attentionmaximizationthrm})).  The NIAC condition~(\ref{eqn:attentionmaximizationthrm}) implies that the cumulative expected utility across all $K$ decision problems is maximized by employing action selection policy $\pi_k(a|x)$ to the $k^{th}$ decision problem. This is identical to the allocation problem in~\cite{KM57} with $A$ defined above as the profitability matrix and the optimal allocation function mapping plants to locations being the identity map. \cite{KM57} solve the allocation problem by defining an equivalent linear program. From duality theory~(\cite{BS97}, Chap.\ $5$), there exist shadow prices $C_k \in \mathbb{R}^{+}$ for each attention selection policy $\pi_k(a|x)$ such that 
 \begin{equation*}
 \begin{split}
 &G_{j,i}-G_{j,j}\leq C_i-C_j \quad \forall i,j\in\{1,2\dots K\}\\
 \implies & G_{i,i} - C_i \leq  G_{i,j} - C_j
 \end{split}
 \end{equation*}
 Note that under the parsimonious representation assumption in Sec.\ \ref{sec:MutualInformationMeasuresandAttentionCost}, the attention function $\alpha_i$ can be replaced with the action selection policy $\pi_i$ defined in~(\ref{eqn:ActionSelection}). Using the shadow prices $C_i$, we construct the cost function $C$ defined in~(\ref{eqn:utilitymaximization}) as follows
\begin{equation}
C(\mu,\alpha)  = 
\begin{dcases}
C_i, & \exists i\in\{1,2\dots K\} \text{ s.t } \alpha=\alpha_i\\
\infty & \text{otherwise}
\end{dcases}
\end{equation}
Clearly, the above construction of the information cost function $C(\mu,\alpha)$ implies that the attention selection policies are optimally chosen over decision problems, hence proving~(\ref{eqn:attentionmaximization}) in Definition~\ref{def:rationalinattention}. 
 %To Include barrier penalty method; and under parsimonious representation, there is a one-to-one mapping between signals and actions. And mention shadow costs method to reconstruct costs for observed attention functions. 
 
 \subsection{Theorem~\ref{thrm:reynicost} }
 \label{appdx:proof_KKT}
In the context of the convex optimization problem defined in~(\ref{eqn:shannoninformationmax}) (\cite{BD04}, pg. $121$), the following KKT conditions are necessary and sufficient for the global optimum $p^{\ast}(x,a)$ to satisfy, provided strong duality (for eg. Slater's condition) holds:
  %\cite{BD04}, pg. $121$ give necessary and sufficient conditions known as KKT conditions for the solution of a convex optimization problem provided strong duality (eg. Slater's condition) holds.  
  %by the solution of the primal and dual solutions $p^{\ast}(x,a)$ and $\lambda_1^{\ast},\lambda_2^{\ast}$ respectively of the convex optimization problem defined in~(\ref{eqn:shannoninformationmax}), provided strong duality (eg. Slater's condition) holds i.e.,
\begin{equation*}
\begin{split}
& \exists \lambda_1^{\ast},\lambda_2^{\ast}\in \mathbb{R} ~s.t\\
 &\nabla \left(\sum_{\mathcal{X},\mathcal{A}}  p^{\ast}(x,a)u(x,a) \right) - \lambda_1^{\ast} \nabla( I_{\beta}(\mu,\alpha^{\ast})-\kappa_{max}) + \lambda_2^{\ast} \nabla(\mu - [p^{\ast}(x)]_{x\in\mathcal{X}} ) = 0\\
& (\mu - [p^{\ast}(x)]_{x\in\mathcal{X}} )=0, \quad  I_{\beta}(\mu,\alpha^{\ast})\leq\kappa_{max}\\
& \lambda_1^{\ast}\geq 0,\quad \lambda_1^{\ast}\left(I_{\beta}(\mu,\alpha^{\ast})-\kappa_{max}\right)= 0.
\end{split}
\end{equation*}
%where $p^{\ast}(x,a)$ denotes the global optimum, $\lambda_1^{\ast},\lambda_2^{\ast} \in \mathbb{R}$.\newline 
Slater's condition can be checked by setting $p(x)=\mu(x), p(a|x)=p_{arb}(a),$ where $p_{arb}(a)$ is any arbitrary probability mass function over the set of actions $\mathcal{A}$. This particular choice of the joint probability mass function $p(x,a)$ results in $I_{\beta}(\mu,\alpha)=0<\kappa_{max},\forall \beta\in (0,1]$ thus satisfying the Slater's condition. 

Let $p^{\ast}_{nc}(x,a)\in \arg\max_{p(x,a)}\left(\sum_{\mathcal{X},\mathcal{A}}  p^{\ast}(x,a)u(x,a) \right)$ denote the unconstrained optimum for the convex optimization problem~(\ref{eqn:shannoninformationmax}) and denote its corresponding attention function (defined in~(\ref{eqn:utilitymaximization})) to be $\alpha_{un}^{\ast}$. Then, setting $\kappa_{max}<I_{\beta}(\mu,\alpha^{\ast}_{un})$ implies $\lambda_1^{\ast}>0$, where $\lambda_1^{ast}$ is the Lagrange multiplier corresponding to the inequality constraint $I_{\beta}(\mu,\alpha)\leq \kappa_{max}$. 
%\begin{equation*}
%\begin{split}
%&\nabla \left(\sum_{\mathcal{X},\mathcal{A}}  p^{\ast}(x,a)u(x,a) \right) - \lambda_1^{\ast} \nabla( I_{\beta}(\mu,\alpha^{\ast})-\kappa_{max}) + \lambda_2^{\ast} \nabla(\mu - [p^{\ast}(x)]_{x\in\mathcal{X}} ) = 0\\
%& (\mu - [p^{\ast}(x)]_{x\in\mathcal{X}} )=0, \quad  I_{\beta}(\mu,\alpha^{\ast})\leq\kappa_{max}\\
%& \lambda_1^{\ast}> 0,\quad I_{\beta}(\mu,\alpha^{\ast}) = \kappa_{max} .
%\end{split}
%\end{equation*}
	Define the Lagrangian $L(p) =  \left(\sum_{\mathcal{X},\mathcal{A}}  p(x,a)u(x,a) \right) - \lambda_1^{\ast} ( I_{\beta}(\mu,\alpha)-\kappa_{max}) + \lambda_2^{\ast} (\mu - [p(x)]_{x\in\mathcal{X}} ) $.
\begin{compactenum}
	\item Consider the Renyi mutual information defined in~(\ref{eqn:renyimutualinformation}) with $\beta \in (0,1)$:
 	\begin{equation*}
 	\begin{split}
 	%&p(x,a) \in \arg\max_{p(x,a)}L(p(.,.)) =  \arg\max_{p(x,a)}\{  \sum_{\mathcal{X},\mathcal{A}}  p(x,a)u(x,a)  - \lambda_1 ( I_{\beta}(\mu,\alpha)) + \lambda_2 (\mu - [p(x)]_{x\in\mathcal{X}} )\}\\
 	%& \text{KKT Conditions:}\\
 	&\frac{\partial}{\partial p(\hat x,\hat a)}  (L(p)) = u(\hat x, \hat a) -\lambda_1^{\ast}\frac{\partial}{\partial p(\hat x,\hat a)} (I_{\beta}(\mu,\alpha))+ \lambda_2^{\ast} = 0\\
 	&\frac{\partial}{\partial p(\hat x,\hat a)}(I_{\beta}(\mu,\alpha))= \frac{\partial}{\partial p(\hat x,\hat a)}\left[\frac{1}{\beta-1}\operatorname{ln} \left( \sum_{x\in\mathcal{X}}\sum_{a\in\mathcal{A}} \frac{p^{\beta}(x,a)}{\mu^{\beta-1}(x)p^{\beta-1}(a)} \right)\right]  \\
 	&=\frac{1}{(\beta-1) \left( \mathbb{E}\left[ \frac{p(x|a)}{\mu(x)}^{\beta-1}\right] \right)} \left[ -(\beta-1)p^{-\beta}(\hat a)\sum_{x\in\mathcal{X}}\frac{p^{\beta}(x,\hat a)}{\mu^{\beta-1}(x)} + p^{1-\beta}(\hat a)\frac{\beta p^{\beta-1}(\hat x,\hat a)}{\mu^{\beta-1}(\hat x)}\right]\\
 	& = \frac{1}{(\beta-1) \left( \mathbb{E}\left[ \frac{p(x|a)}{\mu(x)}^{\beta-1}\right] \right)} \left[ -(\beta-1)p^{1-\beta}(\hat a)\mathbb{E}_{x}\left[\eta^{\beta-1}(x,\hat a)\right] + \beta p^{1-\beta}(\hat a)\eta^{\beta-1}(\hat x, \hat a)\right]\\
 	&= \frac{\beta \eta^{\beta-1}(\hat x,\hat a)-(\beta-1)\mathbb{E}_{x}\left[\eta^{\beta-1}(x,\hat a)\right] }{(\beta-1) {\left( \mathbb{E}\left[ \eta^{\beta-1}(x,a)\right] \right)}p^{\beta-1}(\hat a)} \\
 	&\therefore \boxed{\frac{\partial}{\partial p(\hat x,\hat a)}  (L(p)) = u(\hat x, \hat a) -\lambda_1^{\ast}  \frac{\beta \eta^{\beta-1}(\hat x,\hat a)-(\beta-1)\mathbb{E}_{x}\left[\eta^{\beta-1}(x,\hat a)\right] }{(\beta-1) {\left( \mathbb{E}\left[ \eta^{\beta-1}(x,a)\right] \right)}p^{\beta-1}(\hat a)}      + \lambda_2^{\ast}}
 	\end{split}
 	%\label{eqn:proof_renyi}
 	\end{equation*}\\
 	\item Consider the Shannon mutual information defined in~(\ref{eqn:renyimutualinformation}) with $\beta=1$: 
 	 	\begin{equation*}
 	\begin{split}
 	%&p(x,a) \in \arg\max_{p(x,a)}L(p(.,.)) =  \arg\max_{p(x,a)}\{  \sum_{\mathcal{X},\mathcal{A}}  p(x,a)u(x,a)  - \lambda_1 ( I(\mu,\alpha)) + \lambda_2 (\mu - [p(x)]_{x\in\mathcal{X}} )\}\\
 	%& \text{KKT Conditions:}\\
 	&\frac{\partial}{\partial p(\hat x,\hat a)} (L(p)) = u(\hat x, \hat a) +\lambda_1^{\ast}\frac{\partial}{\partial p(\hat x,\hat a)}(I(\mu,\alpha))   - \lambda_2^{\ast} = 0\\
 	&\frac{\partial}{\partial p(\hat x,\hat a)}(I(\mu,\alpha)) =\frac{\partial}{\partial p(\hat x,\hat a)} \left(\underbrace{ H(\mu)}_{constant} + \sum_{x\in\mathcal{X}}\sum_{a\in\mathcal{A}} p(x,a)\operatorname{ln}(p(x,a)/p(a))\right) \\
 	&=\frac{\partial}{\partial p(\hat x,\hat a)} \left(\sum_{x\in\mathcal{X},a\in\mathcal{A}} p(x,a)\operatorname{ln}(p(x,a)) - \sum_{a\in\mathcal{A}}p(a)\operatorname{ln}(p(a))\right)\\
 	&=p(\hat x,\hat a).\frac{1}{p(\hat x,\hat a)}+ \operatorname{ln}(p(\hat x,\hat a)) - p(\hat a).\frac{1}{p(\hat a)} - \operatorname{ln}p(\hat a) = \operatorname{ln}p(\hat x|\hat a)\\
 	\therefore & \boxed{\frac{\partial}{\partial p(\hat x,\hat a)}  (L(p)) = u(\hat x, \hat a) -\lambda_1^{\ast}\operatorname{ln} p(\hat x|\hat a)       + \lambda_2^{\ast}}
 	\end{split}
 	%\label{eqn:proof_shannon}
 	\end{equation*}
 \end{compactenum}
 \section{Finite Sample Performance Analysis of the Agent's Action-Selection Policy}
 \label{appdx:Risk-AwareUtilityMaximizationbyOff-PolicyEstimation}
 %Theorem~\ref{thrm:decisiontestrationalinattention} can be used for detecting if an agent satisfies rational inattention.
 % Here we illustrate how an observer can optimally construct a set of action selection policies $\boldsymbol{\pi}(a|x,f)=\{\pi_k(a|x,f)\}_{k=1}^K$ over $K$ decision problems that are consistent with rational inattention that maximizes the agent's expected utility. 
 This appendix gives a finite sample analysis of the agent's action selection policy defined in Sec.\ \ref{sec:ProblemFormulationandRationalInattention}.
 Since the results are somewhat tangential to our main application, we have put this analysis in an appendix.
 In YouTube, given the action selection policy (\ref{eqn:ActionSelection}), the observer (data analyst) can construct an optimal recommender system for agents' commenting behavior that is consistent with the agent's commenting preferences while ensuring the agent's commenting behavior satisfies rational inattention. The construction of the optimal policies $\boldsymbol{\pi}(a|x,f)$ is based on a variance-penalized optimization method using finite sample bounds on the total expected utility (\ref{eqn:totalexpectedutility}). 
 
%Thus  far we have constructed estimates for an agent's utility function and information acquisition  cost by observing the agents behavior.
%Indeed,
Consider the maximum likelihood estimate of the agent's action-selection policy $\hat{\pi}(a|x,f)$ (\ref{eqn:policyprior}).
An important  question related to performance analysis of these estimators  is: How far is the net utility obtained using this estimated policy (based on a finite dataset) compared
to the actual net utility $V(\boldsymbol{\pi}(a|x,f))$ (\ref{eqn:totalexpectedutility}) which uses the true policy $\boldsymbol{\pi}(a|x,f)$? Using an extension of the empirical Bernstein inequality to the space of continuous function classes
 \begin{align}
 &\mathcal{F}_\Pi=\{f_{\pi,k}: \mathcal{X}\times\mathcal{A}_k\times{N}\rightarrow [0,1]\}, \; f_{\pi,k} = M\frac{\pi_k(a|x,f)}{\hat{\pi}_k(a|x,f)}u(x,a,f)=M\bar{u}(\pi_k(a|x,f))
 \label{eqn:functionclass}
 \end{align}
 we can  construct a finite sample bound between the observed net utility $V(\hat{\boldsymbol{\pi}}(a|x,f))$ and an estimate of the net utility $V(\boldsymbol{\pi}(a|x,f))$ for the unobserved policy  $\boldsymbol{\pi}(a|x,f)$. In (\ref{eqn:functionclass}), $M$ is a normalization constant which ensures $f_{\pi,k}\in[0,1]$, $\hat{\pi}_k(a|x,f)$ is the observed policy (\ref{eqn:policyprior}), and $\pi_k(a|x,f)$ is an unobserved policy. By bounding the function class (\ref{eqn:functionclass}) using the uniform covering number and employing the double-sampling method~(\cite{AB09}), Theorem~\ref{thrm:policyaccuracy} results. 
 \begin{theorem}
 	Let $\bar{u}(\pi_k)$ be a random variable with $T_k$ independent and identically distributed (i.i.d.) samples in $\mathcal{D}$. Then with probability $1-\gamma$ the random vector $(a_t,x_t)\sim\pi_k$, for a stochastic hypothesis class $\pi_k\in\Pi$, $T_k \geq 16$, and $\lambda=\sqrt{18\operatorname{ln}(10\mathcal{N}_\infty\{1/T_k,\mathcal{F}_\Pi, 2T_k\}/\gamma)}$, satisfies
 	\begin{equation}
 	V(\pi_k) \leq \hat{V}(\pi_k)+\lambda\sqrt{\frac{\operatorname{Var}[\bar{u}(\pi_k)]}{T_k}}+\frac{15\lambda^2}{18M(T_k-1)}
 	\end{equation}
 	where $\mathcal{N}_\infty\{1/T_k,\mathcal{F}_\Pi, 2T_k\}/\gamma)$ is the uniform covering number. $\qed$
 	\label{thrm:policyaccuracy}
 \end{theorem}
 Theorem~\ref{thrm:policyaccuracy}  provides a probabilistic bound between the estimated net utility $\hat{V}(\pi_k)$ and actual net utility $V(\pi_k)$ that only depends on the dataset $\mathcal{D}$ and the coefficient $\lambda$. Therefore, for constructing the true policy $\boldsymbol{\pi}(a|x,f)$, one would maximize the net utility $\hat{V}(\pi_k)$ while minimizing the variance term with a coefficient $\bar{\lambda} \geq 0$. Note that in Theorem~\ref{thrm:policyaccuracy}, $\lambda$ encodes the entropy of the function class $\mathcal{F}_\Pi$,  which is dependent on the number of samples $T_k$, uniform covering number $\mathcal{N}_\infty\{\cdot\}$, and $\gamma$ which is a measure of the confidence of the estimate. For the function class (\ref{eqn:functionclass}), $\mathcal{N}_\infty\{\cdot\}$ is polynomial in the sample size $T_k$~(\cite{MP09,VC15,Sau72})--this ensures as the sample size increases that $\widehat{V}(\pi_k)\rightarrow V(\pi_k)$. 
 Using Theorem~\ref{thrm:policyaccuracy}, the mixed integer-linear program 
 \begin{align}
 &\boldsymbol{\pi}(a|x,f) \in \operatorname*{arg\, max}_{\pi_k\in\Pi}\left\{\sum_{k=1}^KV(\boldsymbol{\pi}_k(a|x,f))-\bar{\lambda}_k\sqrt{\frac{\operatorname{Var}[u(a,x,f)]}{T_k}}\right\} \nonumber\\
 &\text{ s.t. } \quad \sum_{a\in\mathcal{A}_k}\pi_k(a|x,f) = 1, \quad \pi_k(a|x,f) \geq 0 \nonumber\\
 &  \mathcal{L}(u(a,x,f), \hat{\pi}_k(a|x,f)) \nonumber \\
 &  \mathcal{L}(u(a,x,f), \pi_k(a|x,f))  \quad \forall x\in\mathcal{X}, \forall a\in\mathcal{A}_k, \forall k\in\{1,\dots,K\}, \forall f\in\{1,\dots,N\}\\
 & \text{where $\mathcal{L}(\cdot)$ refers to the linear inequalities~(\ref{eqn:MILPagentutility}) and (\ref{eqn:bigM}) in Corollary~\ref{corl:MILP}}.
 \label{eqn:offpolicymaximization}
 \end{align}
 can be used to construct the optimal policy $\pi_k(a|x,f)$ that maximizes the net utility $V(\boldsymbol{\pi}(a|x,f))$ while ensuring the policy is consistent with rational inattention. The regularization term $\bar{\lambda}_k$ in (\ref{eqn:offpolicymaximization}) balances the maximization of the net utility $V(\boldsymbol{\pi}(a|x,f))$ while accounting for the finite-sample variance associated with estimating $V(\boldsymbol{\pi}(a|x,f))$ for policies $\boldsymbol{\pi}(a|x,f)$ that are different from $\hat{\boldsymbol{\pi}}(a|x,f)$. The lower the  value of $\bar{\lambda}_k$, the more risk-seeking the generated optimal policy. 
 
 As seen, the objective in (\ref{eqn:offpolicymaximization}) is based on the finite-sample bound provided in Theorem~\ref{thrm:policyaccuracy}.
 An important property of (\ref{eqn:offpolicymaximization}) is that it provides a method to construct optimal policy recommendations for agents. Specifically, in a state $x$ and frame $f$ recommendations can be tuned such that the probability of selecting action $a$ is consistent with the optimal policy $\boldsymbol{\pi}(a|x,f)$ to maximize the agent's total expected utility  $V$ in (\ref{eqn:totalexpectedutility}) without impacting the preferences of the agent.
 %%%%%%%%%%%%%%%%%%%%%%
 
\section{Denoising Autoencoder Architecture for YouTube Title and Thumbnail}
\label{appdx:DeepClustering}
%A detailed description of 
Algorithm~\ref{alg:framingassociation} (in the main text ) outlined the steps in the deep embedding method for constructing the preference invariant frames. 
%Here we discuss the details. 
Algorithm~\ref{alg:framingassociation_1} below describes the details of the deep clustering procedure explained in Sec.\ \ref{sec:FrameingInformationDeepLearning}.
The denoising autoencoder is comprised of stacked long short term memory (LSTM) and convolutional neural network (CNN) which are detailed in Sec.\ \ref{subsec:textprocessingYouTubetitle} and Sec.\ \ref{subsec:imageprocessingYouTubethumbnail}. To ensure the denoising autoencoder is robust to variations in the title and thumbnail input (e.g. good generalization performance), we introduce noise into the input training data. Possible methods to introduce noise into the network include using drop-out~(\cite{SHKSS14}) and drop-path~(\cite{HSLSW16}) methods. Here we apply Gaussian noise to the input images and numeric representation of the words, and additionally include drop-out layers in the LSTM and CNN networks. 
\begin{algorithm} % enter the algorithm environment
\caption{Deep Embedded Clustering for Framing Association} % give the algorithm a caption
\label{alg:framingassociation_1} % and a label for \ref{} commands later in the document
\begin{algorithmic} % enter the algorithmic environment
    \REQUIRE Set of framing information $\{f_t\}_{t=1}^T$, number of unique frames $N$, stopping threshold $\delta\in(0,1)$, confidence threshold $\delta_c\in(0,1)$, and updating interval $\zeta$. 
    \HEADER{Pre-train}
    	\STATE Pre-train the denoising autoencoder without any frame association.
    \ENDHEADER
    \HEADER{Initialize}
    	\STATE Initialize the $N$ cluster centers $\Psi^o$ using k-means clustering in the latent space and set $\varepsilon=0$.
    \ENDHEADER
        \HEADER{Deep Clustering}
    	\STATE Train the deep clustering autoencoder and frame association layers. 
    	\STATE $i = 0$
    	\WHILE{$\sum_{t}n^o_t \neq n^i_t \geq T\delta$}
    		\STATE \IF{$ i\mod \zeta == 0$}
    			\STATE Compute all latent points $\{z_t=r(w(f_t))\}_{t=1}^T$
    			\STATE Compute $P$ using (9)
    			\STATE Set $n^o=n^i$
    			\STATE Compute new cluster labels $n^i_t = \operatorname*{arg\,max}_{n\in\{1,\dots,N\}}\{q_{in}\}$.
    		\ELSE
    			\STATE Select mini-batch sample from $\{f_t\}_{t=1}^T$ and update the weights of the autoencoder and frame association layers to minimize the loss (7). 
    		\ENDIF
    		\STATE $i=i+1$
    	\ENDWHILE
    \ENDHEADER
\RETURN Invariant frames $n_t \quad\forall t\in\{1,\dots,T\}$ such that $\operatorname{max}_n\{q_{t,n}\} > \delta_c$. 
\end{algorithmic}
\end{algorithm}
\subsection{Text Processing of the YouTube Title}
\label{subsec:textprocessingYouTubetitle}
The design of autoencoders for text data is challenging as a result of the power-law distribution of words~(\cite{MB53}) and the long-range dependencies (grammars) between words. To address these challenges, we use previously constructed word embeddings to convert the words into a numeric vector. We then  employ a LSTM networks for the encoder and decoder blocks of the autoencoder which focus on text processing. The combination of using word embeddings and LSTMs allows the network to utilize prior knowledge of similar words while simultaneously learning how to cluster similar sentences into a unique frame. 

Prior to transforming the words into their numeric embedding, we apply a lemmatization transformation.  Lemmatization reduces the number of variations of words necessary to consider as it groups all the inflected forms a word into a single base representation. For example, the verb ``to walk'' may appear as ``walk'', ``walked'', ``walks'', ``walking'' which are all converted to ``walk'' via the lemmatization transformation. To perform the lemmatization transformation we use the WordNet lemmatizer~\footnote{https://wordnet.princeton.edu/wordnet/}. The WordNet lemmatizer is comprised of two resources, a set of rules which identify the inflectional endings that can be detached from individual words, and a list of exceptions for irregular word forms. WordNet first checks the exceptions, then remove any inflectional endings from the words. Having performed the lemmatization operation, we now construct numeric vector representations of the words. A popular method to perform this task is to use distributed representations of words (e.g. word embeddings). The distributed representation of words in a vector space are designed such that words with similar semantic meaning have similar latent space representations. Equivalently, words with similar meaning will cluster together in the word embedding space. Two popular word embeddings are the Word2Vec~(\cite{MSCCD13}) and Glove~(\cite{PSM14}) models. For the clustering algorithm we use the Glove embedding that was constructed using over 2 billion tweets and is comprised of over 1.2 million words. The possible dimension of the word embedding space is 25, 50, 100, or 200. Here we use a word embedding dimension of 25. 

Given the word embeddings of the sentence $w(f)$, we use an LSTM encoder-decoder framework to learn latent space representations of the titles~(\cite{Gol16,SVL14,GBC16,Ger17}). To construct the latent space representation of the sentences, we utilize a stacked LSTM architecture. Note that stacked LSTMs are able to capture grammatical information in the title at different scales. It was illustrated in~\cite{Gol16,SVL14} that stacked LSTMs tend to have superior predictive performance compared to single layer LSTMs for natural language processing tasks.

\subsection{Image Processing of the YouTube Thumbnail}
\label{subsec:imageprocessingYouTubethumbnail}
In the denoising autoencoder, image processing is performed using a VGG (Visual Geometry Group) based architecture~(\cite{VD15}). Given the latent space representation $z_t$ from the encoder, the image decoder is used to reconstruct the original input image. To perform this task requires the use of deconvolution and upsampling layers. However, deconvolution layers are not used in CNN autoencoders. Instead a mixture of convolutional and upsampling layers are employed. In the most extreme case, a single upsampling layer can be used to directly reconstruct the images from the latent space as illustrated in~\cite{LSD15}. A commonly used method is to construct multiple transposed convolution (also known as fractionally strided convolutions) layers in combination with upsampling layers. Using the transposed convolution layers instead of the standard convolution layers ensures that ``checkerboard'' artifacts are removed from the decoded image~(\cite{ODO16}). 

\section{User Group Statistics in YouTube Dataset}
\label{apdx:youtube}
\begin{figure}[h]
	\centering
	\vspace{-9pt}
	\includegraphics[width=0.722\textwidth]{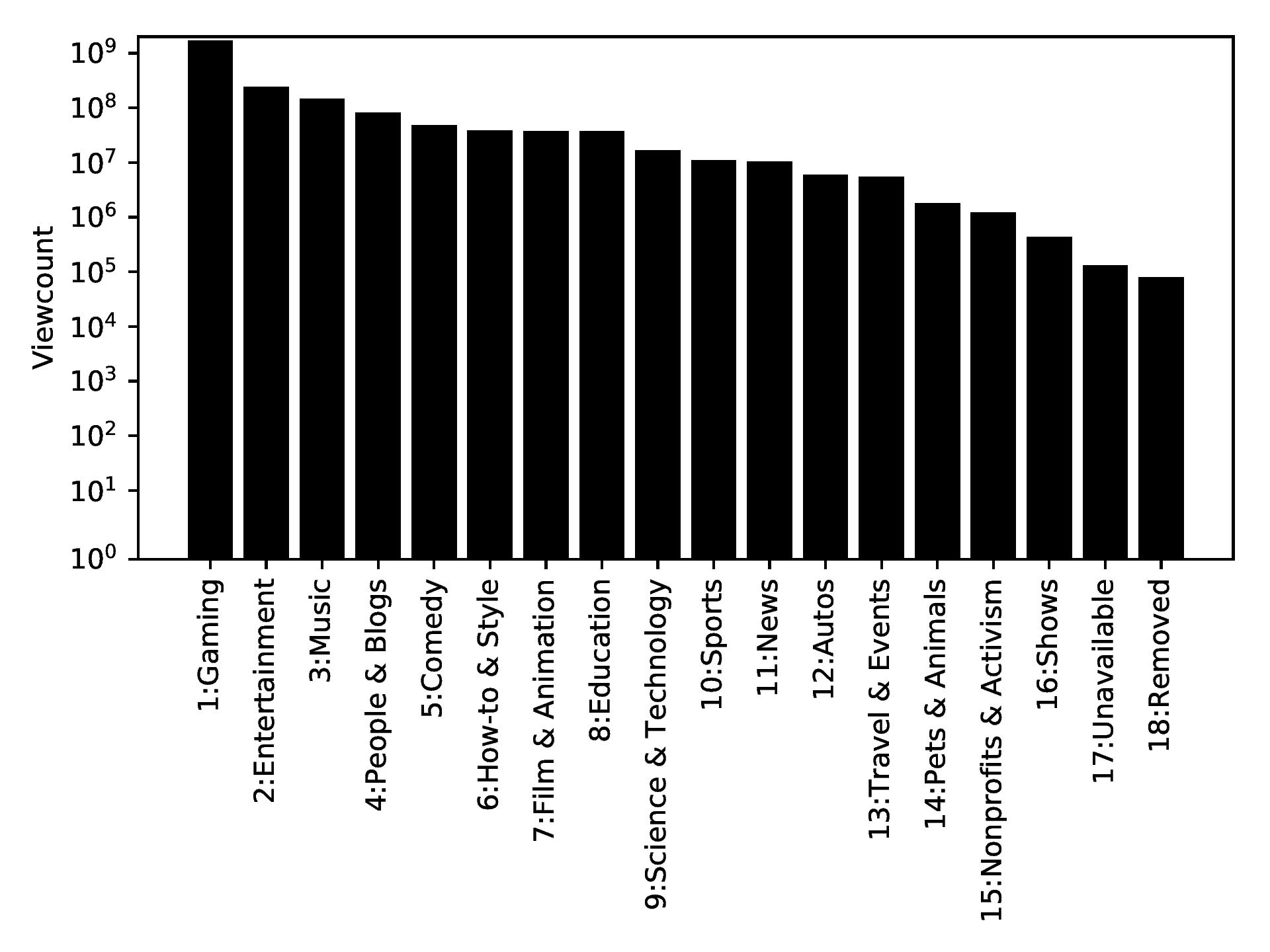}
	\vspace{-10pt}
	\caption{Plot of the viewcount summed over all videos (vertical axis) of the $18$ video categories of the YouTube dataset $\mathcal{D}$. The $18$ categories are listed on the horizontal axis.}
	%(horizontal axis) $k\in\{1,\dots,18\}$ contained in the YouTube dataset $\mathcal{D}$. $k=1$ represents ``Gaming'', $k=2$ ``Entertainment, ..., and $k=18$ ``Removed''.}
	\label{fig:categoryandviewcount}
\end{figure}

In this appendix, we provide additional information about the YouTube dataset analyzed in this paper.
Figure~\ref{fig:categoryandviewcount} lists each video category along with the total number of views. Note that the video categories ``Unavailable'' or ``Removed'' are videos flagged by YouTube as being suspected of violating YouTube's video policies\footnote{Refer to~\url{https://www.youtube.com/yt/about/policies/\#community-guidelines} for details}.

Based on YouTube parameter description in Sec.\ \ref{sec:YTParameters}, each frame (indexed as $1-4$) in Sec.\ \ref{sec:DeepClustering} comprises two decision problems- the first problem for videos in gaming category and the second problem for videos belonging to non-gaming categories. We subdivide the videos in our YouTube dataset into $8$ sub-categories: $1-4$ and $5-8$ correspond to videos belonging to gaming and non-gaming categories in frames $1-4$ respectively.
The average number of interacting users for videos in sub-categories $1-4$ ranges from $493$ to $1133$ users; the average number of interacting users for videos in sub-categories $5-8$ ranges from $368$ to $513$ users.

Similarly in Sec.\ \ref{sec:CatDissect}, each video category (indexed from $1-18$) is associated with a distinct YouTube user group. We again subdivide the videos in our YouTube dataset into $36$ sub-categories where sub-categories $1-18$ and $19-36$ correspond to videos in each video category ($1-18$) with a high viewcount (greater than $10000$ views) and low viewcount (lesser than $10000$ views) respectively. The average number of interacting users for videos in sub-categories $1-18$ ranges from $149$ to $4596$ users; the average number of interacting users for videos in sub-categories $19-36$ ranges from $8$ to $1801$ users.

\section{Estimating the Agent's Attention Function and Choice Function}
\label{apdx:AttentionChoice}
If the dataset $\mathcal{D}$ satisfies rational inattention, the observer (analyst) can estimate the agent's attention function $\alpha_k(s|x)$ and choice function $\eta_k(a|s)$.

Constructing the agent's attention function $\alpha_k(s|x)$ and choice function $\eta_k(a|s)$ which were defined in~(\ref{eqn:utilitymaximization}) and~(\ref{eqn:ActionSelection}) requires the posterior distribution $p_k(x|a)$. First, consider the signal set $\mathcal{S}(\alpha_k)$ of all observed posterior state distributions of the agent for attention function $\alpha_k(s|x)$ using
\begin{equation}
\mathcal{S}(\alpha_k) = \{p_k(x|a): a\in\mathcal{A}_k\}, \quad \label{eqn:revealedposterior}
p_k(x|a) = \frac{\mu(x)\pi_k(a|x)}{\sum_{y\in\mathcal{X}}\mu(y)\pi_k(a|y)}.
\end{equation}
Each posterior distribution $p_k(x|a)$ is associated with a single signal $s\in\mathcal{S}(\alpha_k)$. The posterior distribution $p_k(x|a)$ in (\ref{eqn:revealedposterior}) is equal to the true posterior distribution $p_k(x|s)$ in (1) only if the choice function $\eta_k(a|s)$ produces a single action $a\in\mathcal{A}_k$ for each $s\in\mathcal{S}(\alpha_k)$ with probability one. Otherwise the posterior distribution $p_k(x|a)$ is given by the weighted sum
\begin{equation}
p_k(x|a) = \frac{\sum_{s\in\mathcal{S}(\alpha_k)}\eta_k(a|s)p_k(x|s)p_k(s)}{\sum_{x\in\mathcal{X}}\sum_{s\in\mathcal{S}(\alpha_k)}\eta_k(a|s)p_k(x|s)p_k(s)}.
\end{equation}
Note that without explicit knowledge of the choice and attention functions of the agent, the stochastic choice dataset can not be used to determine if $p_k(x|a)=p_k(x|s)$.  Having $p_k(x|a)=p_k(x|s)$ is not required to determine if the agent satisfies rational inattention. 

Given $p_k(x|a)$, for each signal $s\in\mathcal{S}(\alpha_k)$, the associated attention function is
\begin{equation}
\alpha_k(s|x) = \sum_{a\in\mathcal{A}_k}\eta_k(a|s)\alpha_k(s|x)
=\sum_{a\in\mathcal{A}_k}\pi_k(a|x)\mathbf{1}\{p_k(x|a) = s\}
\label{eqn:attentionfunction}
\end{equation}
where the second equality results from using the data matching expression in~(\ref{eqn:ActionSelection}). Note that (\ref{eqn:attentionfunction}) is only equal to the agent's attention function $\rho_k(r|x)$ if the observed and true posterior distributions are equal. If $\rho_k(r|x)$ is the true attention function then
\begin{align}
\alpha_k(s|x) &= \sum_{r\in\mathcal{S}(\rho_k)}\sum_{a\in\mathcal{A}_k}\eta_k(a|r)\rho_k(r|x)\mathbf{1}\{p_k(x|a) = s\}.
\label{eqn:appdx_attn_selcn}
\end{align}
It must be the case that the observed attention function $\alpha_k(s|x)$ is weakly less informative than the true attention function $\rho_k(r|x)$. Equivalently, the observed attention function is a noisy version of the true attention function. Theorem~1 however does not require we know the true attention function $\rho_k(r|x)$ of the agent to test if the agent's behavior satisfies rational inattention. 

The observed choice function of the agent is given by
\begin{equation}
\eta_k(a|s) = \frac{\sum_{x\in\mathcal{X}}\mu(x)\pi_k(a|x)}{\sum_{b\in\mathcal{A}_k}\sum_{x\in\mathcal{X}}\mu(x)\pi_k(b|x)\mathbf{1}\{p_k(x|b) = s\}}
\label{eqn:choicefunction}
\end{equation}
which is the ratio of the number of times action $a\in\mathcal{A}_k$ was selected over all other possible actions $b\in\mathcal{A}_k$ for the prior distribution $s\in\mathcal{S}(\alpha_k)$. The observed choice function provides no information on the true choice function over the posterior distributions $r\in\Gamma(\rho_k)$ that result from the true attention function unless the actual and observed posterior distributions are equal. Note however that the observed attention function $\alpha_k(s|x)$ (\ref{eqn:attentionfunction}) and choice function $\eta_k(a|s)$ (\ref{eqn:choicefunction}) are consistent with the agent's observed action-selection policy $\pi_k(a|x)$ as expressed in~(\ref{eqn:ActionSelection}).

\bibliographystyle{abbrv}
\bibliography{inattention}
\end{document}